\documentclass[10pt,twocolumn,letterpaper]{article}

\usepackage{geometry}
\usepackage{wacv}
\usepackage{times}
\usepackage{epsfig}
\usepackage{graphicx}
\usepackage{amsmath}
\usepackage{amssymb}
\usepackage{subcaption}
\usepackage[utf8]{inputenc}
\usepackage{lipsum}  
\newcommand{\ra}[1]{\renewcommand{\arraystretch}{#1}}
\usepackage{url}

\wacvfinalcopy % *** Uncomment this line for the final submission

\def\wacvPaperID{393} % *** Enter the WACV Paper ID here

\ifwacvfinal\fi
\begin{document}

%%%%%%%%% TITLE
\title{Does Face Recognition Accuracy Get Better With Age? \\Deep Face Matchers Say No}

\author{Vítor Albiero, Kevin W. Bowyer\\
University of Notre Dame\\
Notre Dame, Indiana\\
{\tt\small \{valbiero, kwb\}@nd.edu}
\and
Kushal Vangara, Michael C. King\\
Florida Institute of Technology\\
Melbourne, Florida\\
{\tt\small \{kvangara2015, michaelking\}@fit.edu}
}

\maketitle
\thispagestyle{empty}

\begin{abstract}
Previous studies generally agree that face recognition accuracy is higher for older persons than for younger persons.
But most previous studies were before the wave of deep learning matchers, and most considered accuracy only in terms of the verification rate for genuine pairs.  
This paper investigates accuracy for age groups 16-29, 30-49 and 50-70, using three modern deep CNN matchers, and considers differences in the impostor and genuine distributions as well as verification rates and ROC curves.  
% In contrast to previous studies,
% that showed face recognition accuracy was better for older persons, 
We find that accuracy is lower for older persons and higher for younger persons.
In contrast, a pre deep learning matcher on the same dataset shows the traditional result of higher accuracy for older persons, although its overall accuracy is much lower than that of the deep learning matchers.
% This contrasts with results for a pre-deep-learning matcher on the same dataset, which gives resultsas used in previous studies whereas our results 
% To explore the causes for this result, we 
Comparing the impostor and genuine distributions, % for the deep learning matchers 
% across the age groups, 
% and also with increasing difference in age between image pairs.  
% Experiments with elapsed time between image pairs show steady or even increasing scores for impostor pairs in the older age group, which is contradictory to what would be expected.  
% Finally, 
we conclude that impostor scores have a larger effect than genuine scores in causing lower accuracy for the older age group.
We also investigate the effects of training data across the age groups.
Our results show that fine-tuning the deep CNN models on additional images of older persons actually lowers accuracy for the older age group.
Also, we fine-tune and train from scratch two models using age-balanced training datasets, and these results also show lower accuracy for older age group.
These results argue that the lower accuracy for the older age group is not due to imbalance in the original training data.
%Promoting the context of reproducible research, the matchers and dataset used in this work are available to the research community.
\end{abstract}

\section{Introduction}
Differences in face recognition accuracy across demographic groups have attracted a lot of attention in recent years.
% Accuracy variations based on race and gender have attracted the most attention.
In this paper, our goal is to develop a better understanding of how face recognition accuracy varies across young, middle and older age groups.
% The literature generally agrees that older persons are easier to recognize and younger are harder.
% However, our results with modern deep CNN face matchers show the opposite comparison.
%

Our experiments use a large, publicly available dataset \cite{morph_site}, that was originally assembled for studying face aging \cite{morph}. 
Importantly, {\it this is the largest generally available dataset that has meta-data for subject age with each image}.
Web-scraped datasets typically do not have such meta-data.
% Our experiments focus on three age ranges: young (16 to 29); middle (30 to 49); and old (50 to 70). 
%
% For each age group, we investigate how genuine scores change with the increase of elapsed time between pairs, and how impostor scores change with the increase of age difference between pairs.

Contributions of this work include:
(1) finding a consistent qualitative effect across race,
(2) investigation of how genuine and impostor distributions contribute to accuracy across age groups,
(3) investigation of the effect of elapsed time between genuine pairs, 
(4) insights on how the effect of age difference between persons in an impostor pair depends on the age groups,
(5) investigation of the effects of balancing the training data across age groups, and
(6) use of publicly available matchers and dataset, supporting reproducible experiments.

%The rest of this paper is organized as follows. Section \ref{sec:related_work} summarizes the previous studies; Section \ref{sec:dataset} describes the dataset, the curation process, and the age groups creation; Section \ref{sec:matchers} describes the face matchers used; Section \ref{sec:results} discuss the experimental results; 
%Section \ref{sec:conclusion} concludes; and finally, Section \ref{sec:discussion} presents discussion points.

\section{Related Work}
\label{sec:related_work}
The 2002 Face Recognition Vendor Test \cite{frvt} looked at how identification rate varies with age for three commercial matchers, and concluded that older persons are recognized more accurately than younger.
Beveridge et al. \cite{Beveridge2009}, using the Face Recognition Grand Challenge dataset \cite{frgc}, found that verification rate increases as subject age increases.
% , and concluded that older persons are recognized more accurately.
In a 2009 meta-analysis of previous works, 
Lui et al. \cite{Lui2009} found that 20 out of 22 age-related results agree that the accuracy is higher for older persons. 
In a 2018 review, Abdurrahim et al. \cite{Abdurrahim2018}  found general agreement in the literature that older persons are recognized more accurately.

% More similar to our work, 
Klare et al. \cite{Klare2012} evaluated six matchers on the age groups 18-30, 30-50 and 50-70, using the Pinellas County Sheriff’s Office (PCSO) dataset. 
The three commercial matchers showed lower verification rates for 18 to 30 age group, indicating that younger people are harder to verify.
%Also, despite not agreeing on whether the middle or old age group has the highest accuracy,
Also, despite not agreeing on which age group has the highest accuracy, 
the two non-trainable matchers (Local Binary Patterns and Gabor filters) show an overall lower accuracy for the young age group.
%
% do we need the reference below?  It doees not do age ranges, right?
%
Best-Rowden and Jain \cite{longitudinal_c} reported results for two commercial matchers on a subset of the PCSO dataset, separated to study the effects of elapsed time between genuine pairs. 
% In their work, t
% They used the first image from each subject as enrollment, and then match the later images, increasing the elapsed time between them.
% From their experiments, the authors 
They conclude that the genuine scores significantly decrease over time for both matchers.
An extended version of their work \cite{longitudinal_j} added an additional dataset and more matchers. 
% They conducted similar studies that show a similar decrease in the genuine score as the elapsed time between enrollment and query images increases.

In 
% a presentation on 
the FRVT Ongoing effort at NIST, Grother \cite{grother} reported higher false match rates (FMRs) for older persons than for younger. 
With a fixed threshold to give an overall FMR of 1-in-10,000, a group of age 20-somethings had a FMR of 1-in-100, but a group with 70-somethings had a FMR of 5-in-100.
On the other hand, they reported higher false non-match rates (FNMRs) for younger subjects, 0.05 for 20-somethings, and 0.02 for 70-somethings.

Lu et al. \cite{Lu2018} looked at results from four deep learning matchers for the IARPA JANUS Benchmark B (IJB-B) dataset, which is a collection of web-scraped images. 
The authors found that accuracy increases with age up until age  50, which agrees with previous results in the literature.
They also found that after age 50, accuracy starts to drop.
However, the age annotations for IJB-B images were derived using Amazon Mechanical Turk, and we are not aware of how accurate they may be.

Cook et al. \cite{cook2018} analyzed  demographic factors in the 2018 Biometric Technology Rally \cite{bio_rally}.
% , with emphasis on race, which was measured using skin reflectance. 
They used images captured from eleven different image acquisition systems, and matched them against same-day enrollment images and against historical images using a  commercial matcher. 
Their age experiment was conducted using two age groups: 20 to 40; and 40 to 85.
% , which were both fine-grained using gender and skin reflectance.
While the same-day matching shows a similar result for both age ranges, they report higher accuracy for older persons in the historical matching.  
%This result suggests that older persons have a more stable face over time than younger people, which leads to a better performance while matching faces with bigger elapsed time between them.

Cao et al. \cite{vggface2} analyzed results of matching young (less than 34 years) and mature (34 years or more) persons, using a deep CNN matcher.
They report that matching young-to-mature faces is harder than young-to-young or mature-to-mature.
They also report that mature-to-mature matching results in higher similarity scores than young-to-young matching.

Previous works generally focus on how match scores for genuine pairs vary across age groups \cite{Beveridge2009, cook2018, vggface2}, and do not explicitly consider both the impostor and genuine distributions.
%or false match rates (FMRs) and false non-match rates (FNMRs).
Some previous works use datasets that are not available 
%to the research community
\cite{Klare2012, grother, cook2018}.
Further, most previous work does not experiment with race or gender subsets, which is important as an imbalance in other demographics across age groups could be a confounding factor for accuracy variations.
% Table \ref{tab:related_work_table} summarizes elements of various previous studies.

We present the most extensive analysis to date of how face recognition accuracy varies across age groups. 
In addition to ROC curves, we analyze genuine/impostor score distributions,
% and compare FMR/FNMR, 
and how they contribute to accuracy differences.
We analyze how elapsed time between genuine pairs and age difference between impostor pairs affects accuracy.
We also investigate how training data affects 
accuracy,
% the performance across age groups, 
with models fine-tuned across age groups, age balanced datasets, and trained from scratch on age balanced datasets.

\section{Dataset}
\label{sec:dataset}

We identified the largest dataset that is available to the research community and that has meta-data for subject age at the time each image was acquired \cite{morph}. 
Many well-known datasets do not have the age of the subject recorded with each image.
This is the case for the IJB datasets \cite{ijbb, ijbc}, MegaFace \cite{megaface}, and more generally, for any web-scraped dataset.
% that we are aware of.
Some datasets used in previous studies are not available to others.
% to the research community \cite{Klare2012,grother,cook2018}.
FG-NET \cite{fgnet} has images with annotated ages, however the number of images with a subject older than 50 years is only 23.
In contrast, the MORPH \cite{morph} dataset used in this work has recorded age meta-data and a relatively large number of subjects and images across the age groups, as it was  collected to study face aging.

The MORPH dataset \cite{morph} was assembled from public records and has been widely used in face aging research. 
We curated a subset of MORPH Album 2, 
%In the first steps of curation, we corrected 1\% of the dataset:
%we deleted 259 images that do not contain a face; deleted 140 images that were repeated; corrected 18 images with wrong race; and corrected 3 images with wrong gender.
%After initial curation, the dataset has 
which, after removal of a small number of images that do not contain a face or that are repeated images, has 53,231 images of 13,119 subjects.
For each image, the meta-data includes a date of birth (DOB) and date of acquisition. 
This makes it possible to split the dataset into images of subjects for different age ranges.
The age ranges used are 16 to 29, 30 to 49, and 50 to 70, termed “young”, “middle” and “old”, respectively.

In initial studies, we found that 1,708 of the 13,119 subjects had inconsistent DOBs across their images. 
As extreme cases, one subject had six different DOBs across their images, and 
another had a 32-year difference across the DOBs for their images. 
However, in the large majority of cases, the inconsistent DOBs could be corrected in a straightforward manner. 
If the DOB was consistent for 75\% or more of a subject’s images, then the inconsistent DOBs for the subject’s other images were corrected to the consistent DOB.
This resolved DOB inconsistency for 1,364 subjects.
%Many of the remaining 798 subjects had a common year for  the DOB across their images, but had varying month or day entries.
%For subjects with the same year in the DOB for 75\% or greater of their images, but varying month or day, we took the median date across the images with consistent year, and corrected the DOB for all additional images.
%This resolved another 454 subjects with inconsistent DOB.
The DOB for the remaining 344 was deemed too inconsistent to correct, and these subjects were dropped. 
Thus the total number of subjects in the dataset for our analysis is 12,775.
The image names and meta-data corrections used in this paper will be made available online.

\begin{table}[t]
    \small
    \setlength\tabcolsep{2pt}
    \centering
    \ra{1.1}
    \begin{tabular}{l|rr|rr|rr}
        \multicolumn{1}{l|}{} & \multicolumn{2}{c|}{\textbf{Whole Dataset}} & \multicolumn{2}{c|}{\textbf{AA Male}}& \multicolumn{2}{c}{\textbf{C Male}}\\
        \multicolumn{1}{l|}{\textbf{Group}} & \multicolumn{1}{c}{\textbf{Subjects}} & \multicolumn{1}{c|}{\textbf{Images}} & \multicolumn{1}{c}{\textbf{Subjects}} & \multicolumn{1}{c|}{\textbf{Images}} & \multicolumn{1}{c}{\textbf{Subjects}} & \multicolumn{1}{c}{\textbf{Images}} \\ \hline
        \textbf{Young}& 5,778 & 21,665 & 4,085 & 16,799 & 833 & 2,690 \\
        \textbf{Middle} & 6,532 & 25,604 & 4,235 & 16,891 & 1,140 & 4,140 \\
        \textbf{Old}& 1,074 & 3,622& 726 & 2,387& 221 & 837
    \end{tabular}
    \caption{Division of age groups for whole dataset, African American (AA) males and Caucasian (C) males.}
    \label{tab:dataset_groups}
    \vspace{-1em}
\end{table}
\begin{figure*}[t]
    \centering
    \begin{subfigure}[b]{0.28\textwidth}
      \centering
      \begin{subfigure}[b]{.4\columnwidth}
        \centering
        \includegraphics[width=\linewidth]{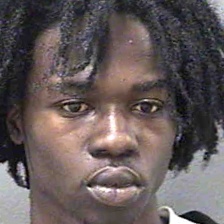}
      \end{subfigure}
      \begin{subfigure}[b]{.4\columnwidth}
        \centering
        \includegraphics[width=\linewidth]{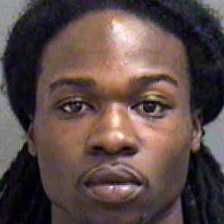}
      \end{subfigure}
      \caption{22 and 25 years}
      \vspace{-0.5em}
  \end{subfigure}
  \begin{subfigure}[b]{0.28\textwidth}
      \centering
      \begin{subfigure}[b]{.4\columnwidth}
        \centering
        \includegraphics[width=\linewidth]{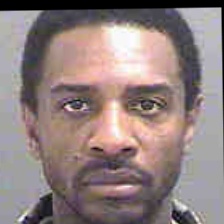}
      \end{subfigure}
      \begin{subfigure}[b]{.4\columnwidth}
        \centering
        \includegraphics[width=\linewidth]{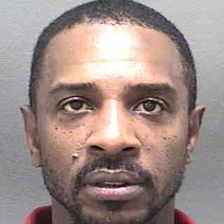}
      \end{subfigure}
      \caption{41 and 44 years}
      \vspace{-0.5em}
  \end{subfigure}
  \begin{subfigure}[b]{0.28\textwidth}
      \centering
      \begin{subfigure}[b]{.4\columnwidth}
        \centering
        \includegraphics[width=\linewidth]{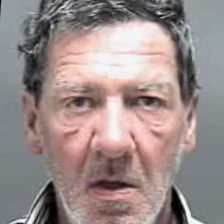}
      \end{subfigure}
      \begin{subfigure}[b]{.4\columnwidth}
        \centering
        \includegraphics[width=\linewidth]{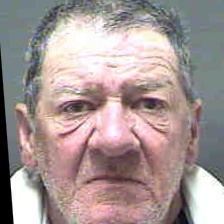}
      \end{subfigure}
      \caption{58 and 61 years}
      \vspace{-0.5em}
  \end{subfigure}
  \caption{Sample authentic pairs with 3 years difference in young, middle and old groups.}
  \label{fig:sample_faces}
  \vspace{-1em}
\end{figure*}

The dataset with curated DOB was divided into the three age ranges.
Note that a particular subject may have images in both the “young” and “middle” range, or the “middle” and “old” range.
However, if a subject has only one image for a given age range, the subject and image were dropped from that age range.
The numbers of subjects and images for the division of the whole dataset into young, middle and old age ranges is given in Table \ref{tab:dataset_groups}.

Other research has shown that face recognition accuracy 
is different for African-American and Caucasian, and for male and female \cite{Klare2012, Beveridge2009, cook2018, afrbp1}.
The young, middle and old ranges of our dataset may have varying demographic mix.
To create subsets with consistent demographic mix across age ranges, we split the dataset into four subsets: African-American male, African-American female, Caucasian male and Caucasian female. 
However, the female old age range had 
% a small number of subjects and images in the old group, 
just 93 subjects and 286 images for African-American, and 34 subjects and 112 images for Caucasian.
This is too few subjects and images for reliable analysis, and so the subgroup analysis was done only for the two male subsets.
Sample authentic pairs are shown in Figure \ref{fig:sample_faces}.
Results of our analysis are presented for the whole dataset, with its varying demographics across age ranges, and for the African-American male and the Caucasian male subsets.

\section{Deep Learning Face Matchers}
\label{sec:matchers}

We use three recent deep CNN face matchers. These matchers are chosen to represent training with different loss functions and different training sets: (1) VGGFace2 (ResNet-50) \cite{resnet}, trained on VGGFace2 dataset \cite{vggface2} with standard softmax loss; (2) FaceNet \cite{facenet}, trained on MS-Celeb-1M dataset \cite{ms1_celeb} with triplet loss; and (3) ArcFace \cite{arcface} (ResNet-100), trained on MS-Celeb-1M V2 dataset with additive angular margin loss.
%which combines the margin penalization from SphereFace \cite{sphereface}, ArcFace and CosFace \cite{cosface}.
Each network was used with pre-trained weights that are publicly available \cite{keras_vggface, keras_facenet, insightface}.

As input to the matchers, the faces were detected and aligned using MTCNN \cite{mtcnn}, and resized to 224x224 pixels (VGGFace2), 160x160 (FaceNet), or 112x112 (ArcFace). 
%The alignment consisted of centering the eyes in the image, and if the face crop had less than 224 pixels, the background pixels were filled with black pixels.
%In a few cases no face was detected (113 images) or the face was aligned improperly (30 images). 
%In such cases, the face was manually aligned, cropped and resized. 
% Figure \ref{fig:mean_faces} shows the “average face” for the different age ranges, computed after face detection, cropping, resizing and alignment.
%
For feature extraction, the last but one layer was used, which corresponds to a 2048-d feature vector for VGGFace2, 128-d for FaceNet, and 512-d for ArcFace.
After extraction, the features were matched using cosine similarity.
\begin{figure*}[t]
    \centering
    \begin{subfigure}[b]{0.8\textwidth}
        \centering
        \begin{subfigure}[b]{0.327\textwidth}
            \centering
            \includegraphics[width=1\columnwidth]{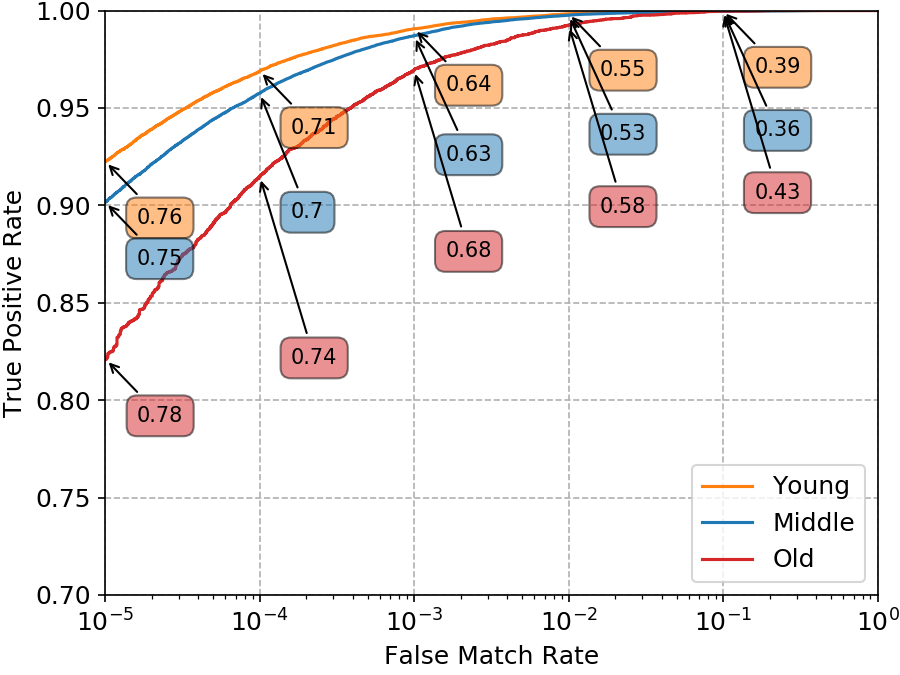}
        \end{subfigure}
        \hfill 
        \begin{subfigure}[b]{0.327\textwidth}
            \centering
            \includegraphics[width=1\columnwidth]{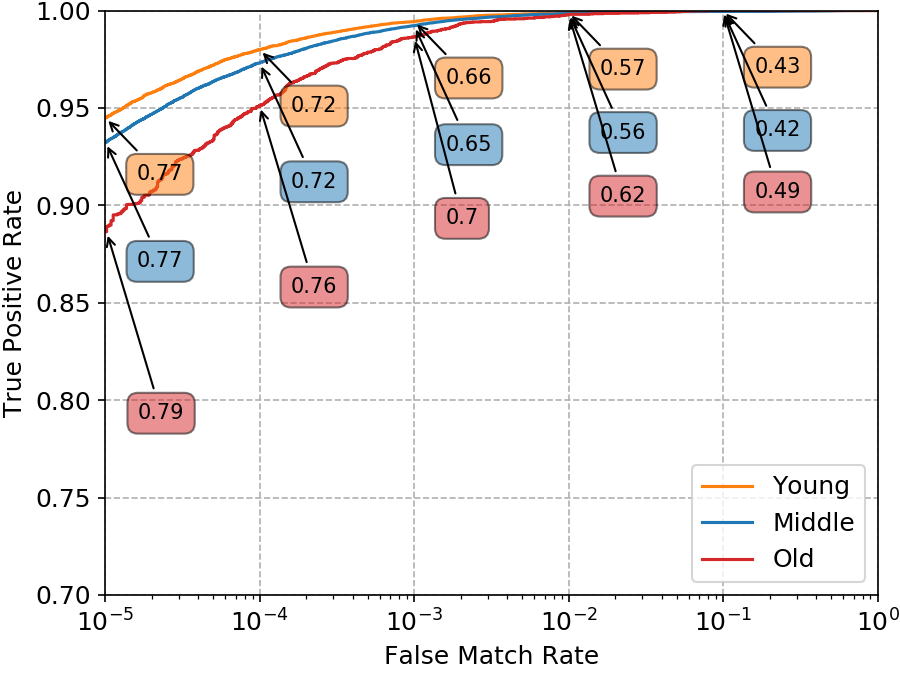}
        \end{subfigure}
        \hfill 
        \begin{subfigure}[b]{0.327\textwidth}
            \centering
            \includegraphics[width=1\columnwidth]{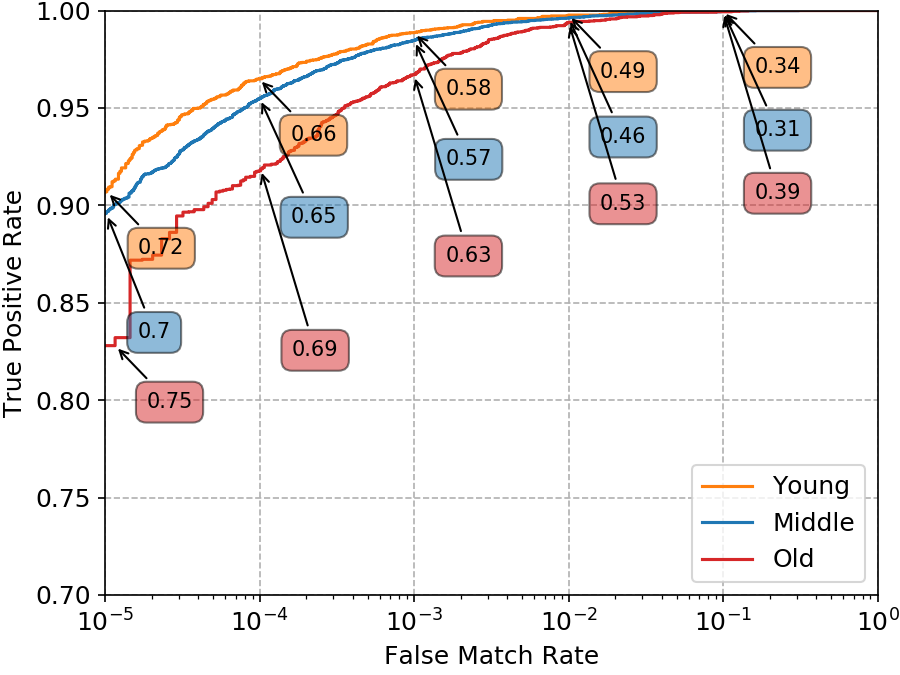}
        \end{subfigure}
    %\end{subfigure}
    \hfill 
    %\begin{subfigure}[b]{1\textwidth}
        \begin{subfigure}[b]{0.327\textwidth}
            \centering
            \includegraphics[width=1\columnwidth]{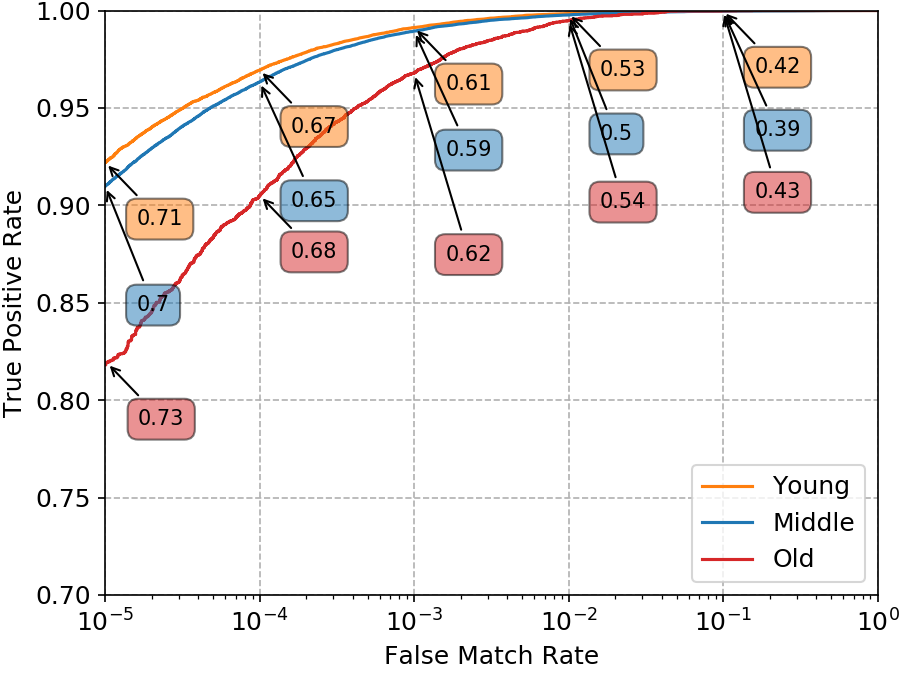}
        \end{subfigure}
        \hfill 
        \begin{subfigure}[b]{0.327\textwidth}
            \centering
            \includegraphics[width=1\columnwidth]{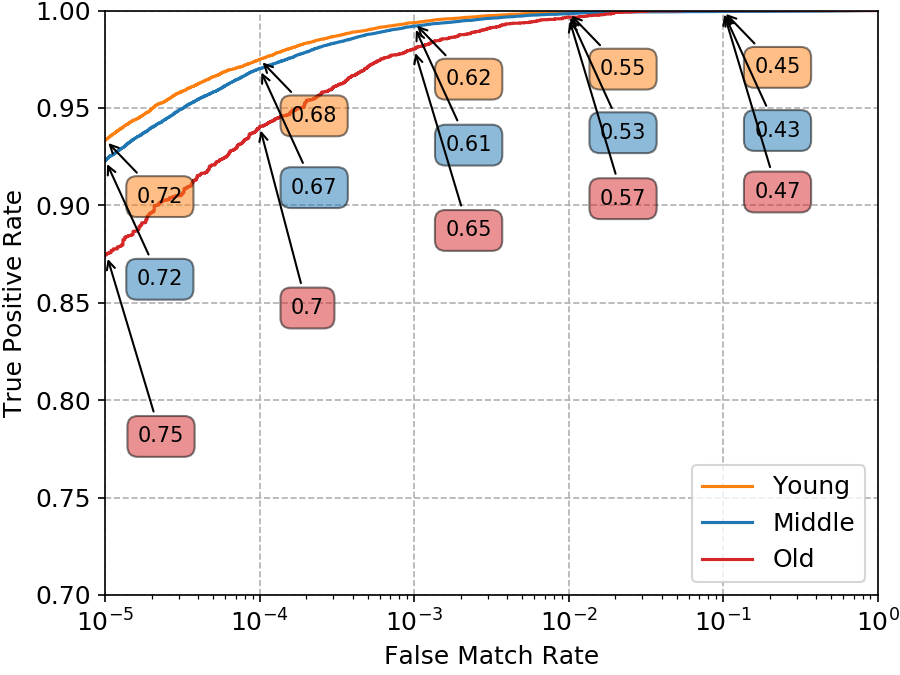}
        \end{subfigure}
        \hfill 
        \begin{subfigure}[b]{0.327\textwidth}
            \centering
            \includegraphics[width=1\columnwidth]{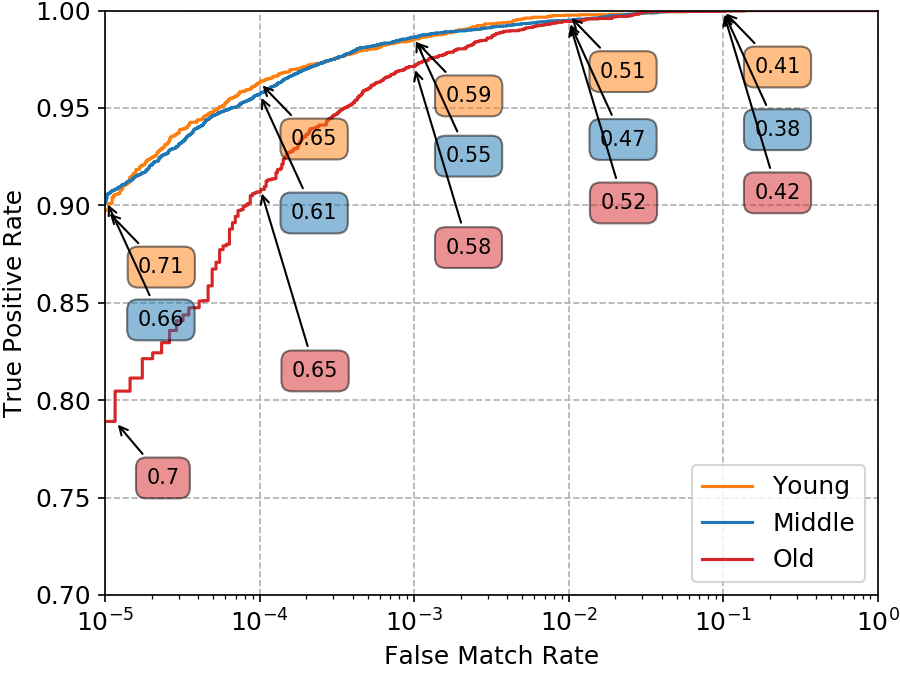}
        \end{subfigure}
        
    \hfill 
    %\begin{subfigure}[b]{1\textwidth}
        \begin{subfigure}[b]{0.327\textwidth}
            \centering
            \includegraphics[width=1\columnwidth]{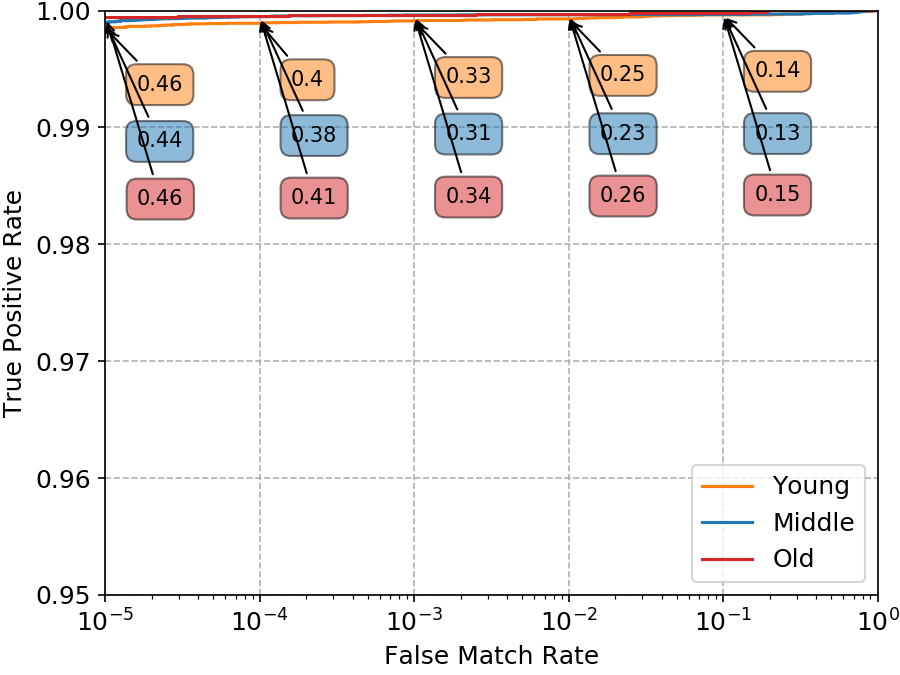}
            \caption{Whole dataset}
            \label{fig:roc_resnet_whole_group}
            \vspace{-0.5em}
        \end{subfigure}
        \hfill 
        \begin{subfigure}[b]{0.327\textwidth}
            \centering
            \includegraphics[width=1\columnwidth]{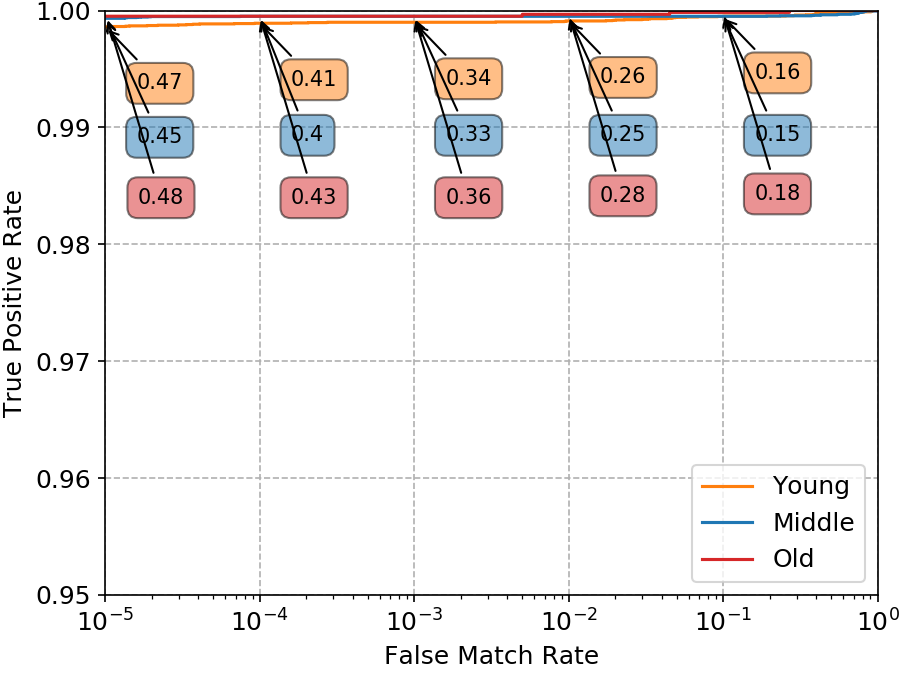}
            \caption{African American males}
            \label{fig:fig:roc_resnet_male_aa}
            \vspace{-0.5em}
        \end{subfigure}
        \hfill 
        \begin{subfigure}[b]{0.327\textwidth}
            \centering
            \includegraphics[width=1\columnwidth]{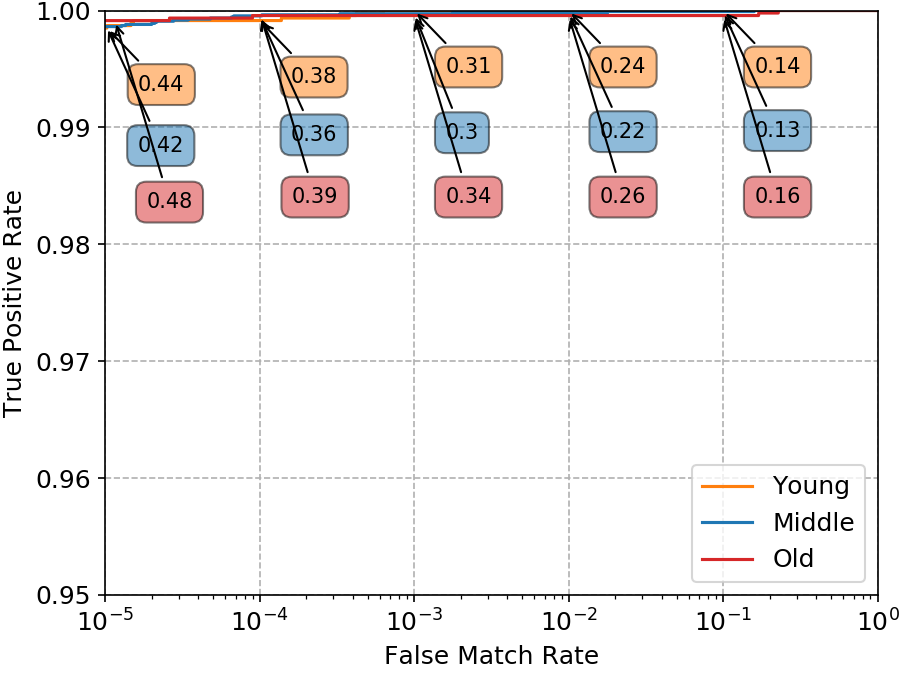}
            \caption{Caucasian males}
            \label{fig:fig:roc_resnet_male_c}
            \vspace{-0.5em}
        \end{subfigure}    
    \end{subfigure}
    \caption{ROC curves for FaceNet (top), VGGFace2 (middle), and ArcFace (bottom). Annotated values correspond to thresholds used for the correspondent FMR. ArcFace is displayed at a different scale for better visualization.}
    \label{fig:roc}
    \vspace{-1em}
\end{figure*}

\section{Experimental Results}
\label{sec:results}
This section first presents ROC results for the matchers, for the whole dataset and for the African-American male and Caucasian male cohorts. 
The relative accuracy shown by the ROC curves across the age ranges is different from what most past studies have found.
%Our results show that the older age group has the worst ROC curve and the younger age range has the best.
To better understand the effects of the two types of errors that are summarized in the ROC curve, we next present the impostor and genuine distributions. 
%, and the FMR and FNMR curves.
We further investigate how elapsed time between impostor and genuine pairs affects their matching scores across the age groups.
%To better understand the differences in the impostor and genuine distributions 
%(and so also the FMR and FNMR curves) 
%of the different age ranges, we consider 
Moreover, to understand why our results differ from those of some previous studies, we present results from a pre deep CNN face matcher as used in a well-known previous work \cite{Klare2012}.
Finally, we analyze how the training data affects the performance of the deep models, by fine-tuning models on separate age groups, age balanced subsets and training from scratch on age balanced subsets.

\subsection{ROC Curve Comparison}
\label{sec:roc_curves}
\begin{figure*}[t]
    \centering
    \begin{subfigure}[b]{0.82\textwidth}
        \centering
        \begin{subfigure}[b]{0.327\textwidth}
            \centering
            \includegraphics[width=1\columnwidth]{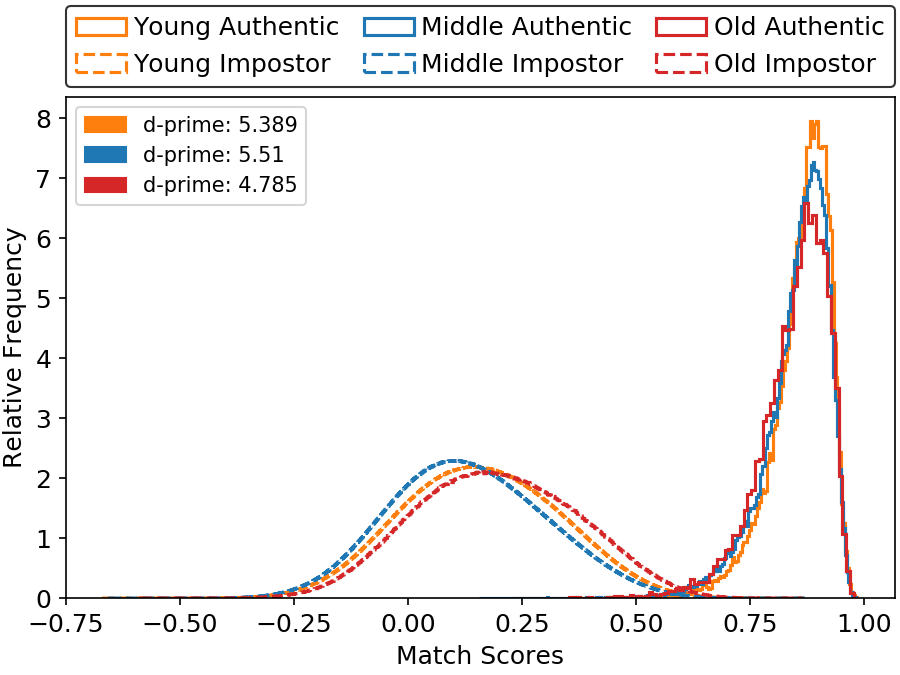}
        \end{subfigure}
        \hfill 
        \begin{subfigure}[b]{0.327\textwidth}
            \centering
            \includegraphics[width=1\columnwidth]{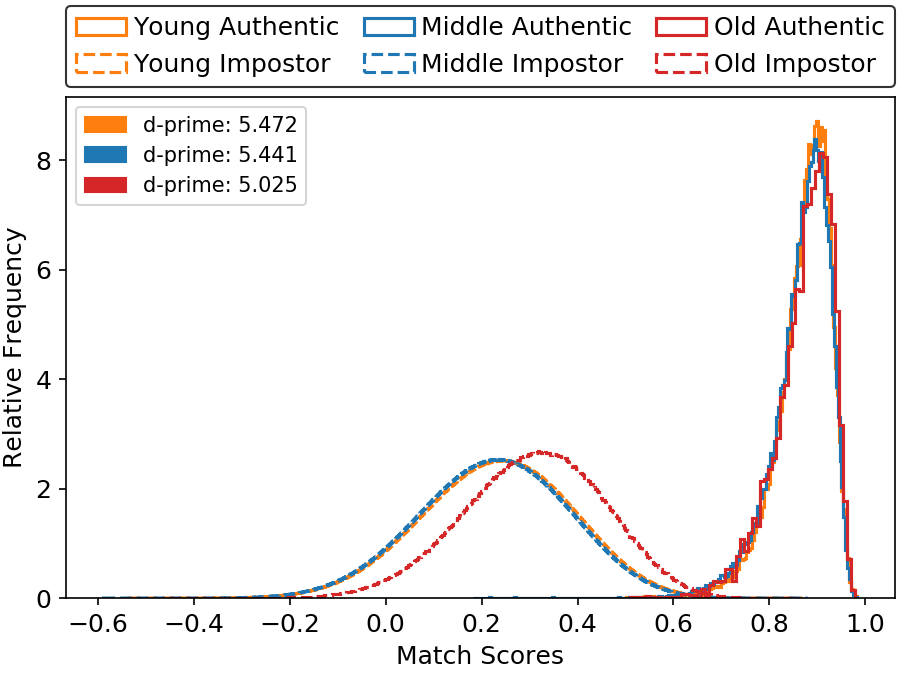}
        \end{subfigure}
        \hfill 
        \begin{subfigure}[b]{0.327\textwidth}
            \centering
            \includegraphics[width=1\columnwidth]{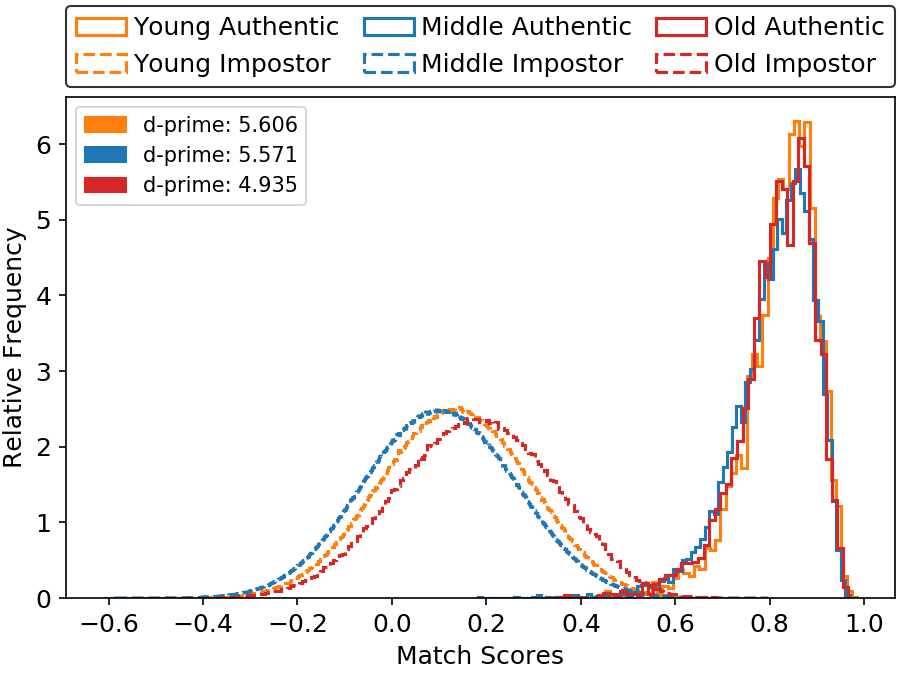}
        \end{subfigure}
    %\end{subfigure}
    \hfill 
    %\begin{subfigure}[b]{1\textwidth}
        \begin{subfigure}[b]{0.327\textwidth}
            \centering
            \includegraphics[width=1\columnwidth]{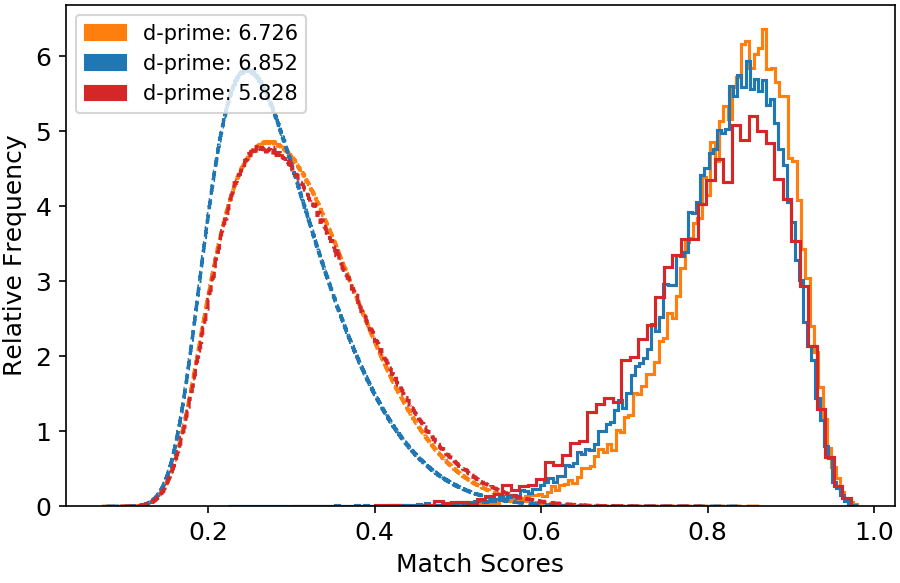}
        \end{subfigure}
        \hfill 
        \begin{subfigure}[b]{0.327\textwidth}
            \centering
            \includegraphics[width=1\columnwidth]{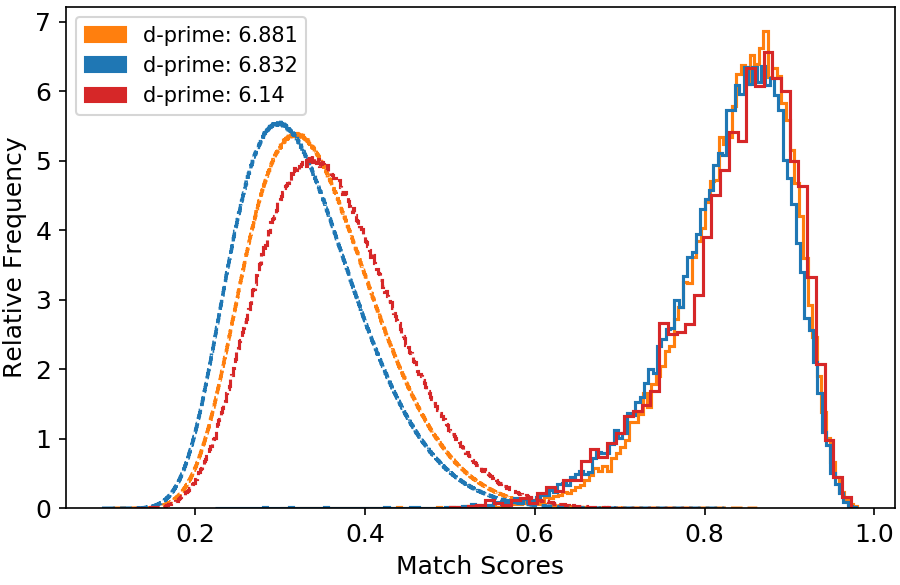}
        \end{subfigure}
        \hfill 
        \begin{subfigure}[b]{0.327\textwidth}
            \centering
            \includegraphics[width=1\columnwidth]{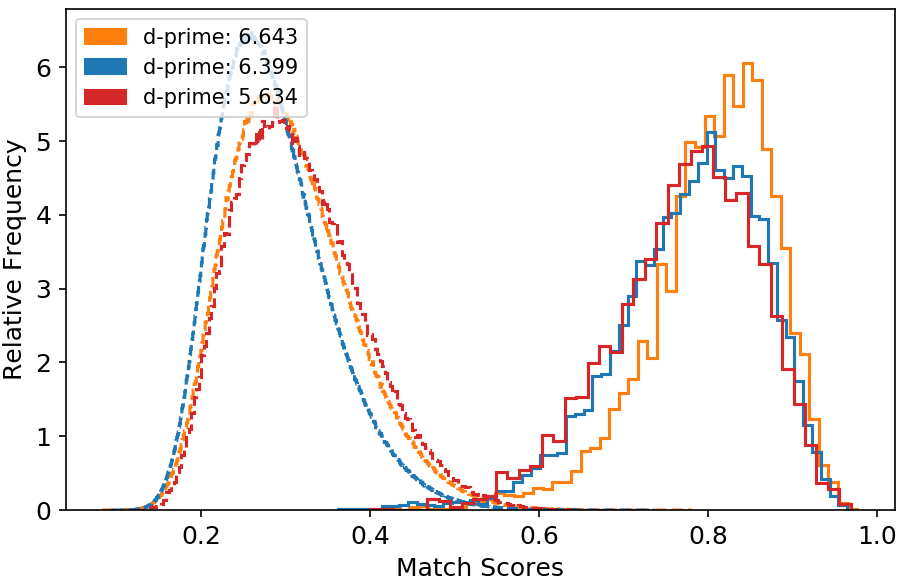}
        \end{subfigure}
        
    \hfill 
    %\begin{subfigure}[b]{1\textwidth}
        \begin{subfigure}[b]{0.327\textwidth}
            \centering
            \includegraphics[width=1\columnwidth]{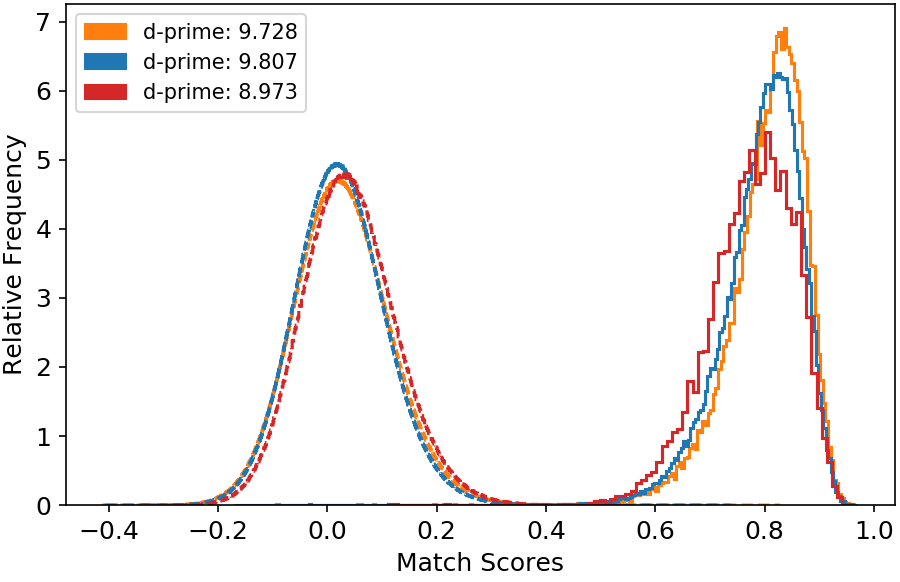}
            \caption{Whole dataset}
            \label{fig:hist_resnet_whole_group}
            \vspace{-0.5em}
        \end{subfigure}
        \hfill 
        \begin{subfigure}[b]{0.327\textwidth}
            \centering
            \includegraphics[width=1\columnwidth]{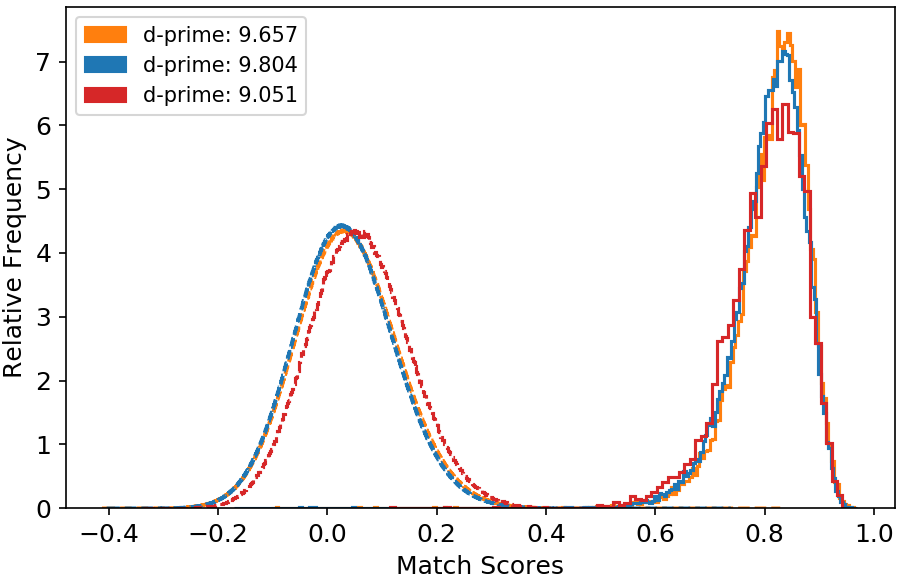}
            \caption{African American males}
            \label{fig:fig:hist_resnet_male_aa}
            \vspace{-0.5em}
        \end{subfigure}
        \hfill 
        \begin{subfigure}[b]{0.327\textwidth}
            \centering
            \includegraphics[width=1\columnwidth]{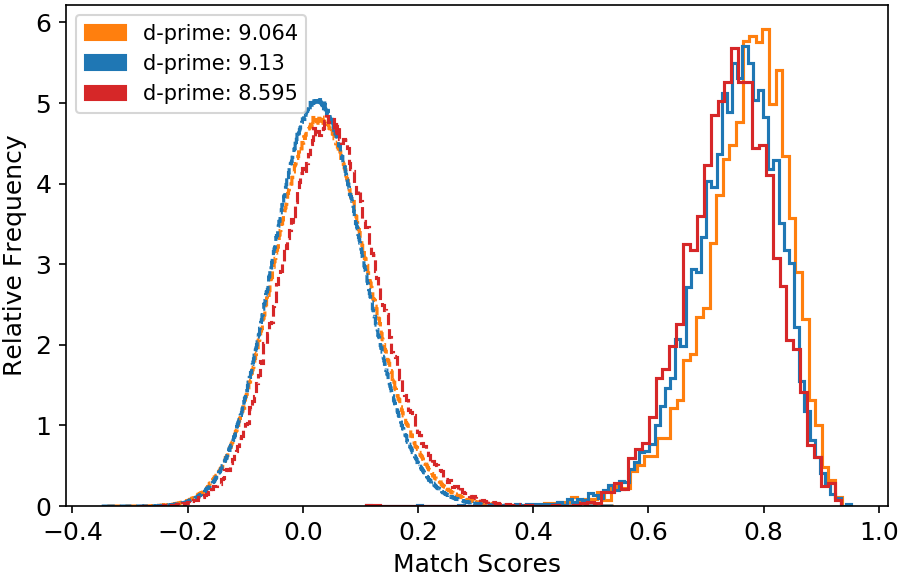}
            \caption{Caucasian males}
            \label{fig:fig:hist_resnet_male_c}
            \vspace{-0.5em}
        \end{subfigure}    
    \end{subfigure}
    \caption{Match scores distribution for FaceNet (top), VGGFace2 (middle), and ArcFace (bottom).}
    \label{fig:hist}
    \vspace{-1em}
\end{figure*}
ROC curves for the different age ranges are compared 
% for the whole dataset and the two subsets, for the three matchers, 
in Figure \ref{fig:roc}.
The general pattern of the ROC results is consistent, whether considering the whole dataset with its varying mixture of gender and race across age ranges, or considering either of the two same-race-same-gender subsets.
It is important to note that ROC curves are more appropriately used to compare accuracy of different algorithms on the same dataset.
In this case, we are comparing the accuracy of the same algorithm for different datasets (different age ranges).
The ROC format hides the fact that the same FMR for different age ranges is obtained at different decision thresholds.
For this reason, the decision threshold value is marked on the plots for sample values of FMR.

We see that except for VGGFace2 on Caucasian males, for all ROC curves, the old age range has higher thresholds to achieve the same FMR.
% as the other groups.
Following the old age range, the young age range has the second highest thresholds.
The thresholds give an idea of how much the overlap of the impostor and genuine distributions is shifted towards higher similarity scores, which can be correlated to a worse impostor distribution.

For six out of nine ROC curves, the young age range has the best ROC, and the old age range has the worst.
%Except for the ArcFace matcher, the middle age range shows better results than the old age range, and the old age range has the worst results.  
For FaceNet and VGGFace2, the gap between middle and old is noticeably larger than the gap between young and middle.
With the ArcFace matcher, the older group has a slightly better ROC curve than the other two age groups, but achieved using a higher threshold.
In the whole dataset, the ArcFace true positive rate (TPR) with a false match rate (FMR) of $10^{-4}$ is 99.89\%, 99.95\% and 99.95\% for the young, middle and old age groups, respectively.
Overall, the pattern of ROC results disagrees with previous studies that found that older persons are easier to recognize than younger persons.
% As mentioned above, the ROC format hides some important information when used to compare accuracy of the same algorithm applied to different datasets.
For this reason, the next section presents the impostor and genuine distributions that underlie the ROC curves.

\subsection{Impostor and Genuine Distributions}
\label{sec:hist}
The impostor and genuine distributions that underlie the ROC curves in Figure \ref{fig:roc} are shown in Figure \ref{fig:hist}.
We see that in each of the nine plots, the impostor distribution for the old group is the worst, with a noticeable shift toward the genuine distributions.
The impostor distribution for the middle age range is the best, with a slight shift toward lower similarity scores.
And the impostor distribution for the young age range is between the other two.

The young age range generally has the best genuine distribution, showing a higher peak of high-similarity scores than the other two age ranges.
There is not a noticeable difference between the genuine distributions for the middle and old age ranges.

From analyzing the impostor and genuine distributions, we can infer that the main factor driving the poorer ROC curves for the old age range is its poorer impostor distribution.
We can also infer that the main factor driving the generally better ROC curves for the young age range is its (slightly) better genuine distribution. 

The impostor and genuine distributions are not perfectly Gaussian.
However, the d-prime statistic may still indicate the relative separation of the two distributions. 
% for the different age ranges.
%For the FaceNet matcher on the whole dataset, the d-prime for the younger, middle and old age range distributions is 5.215, 5.265, and 4.577, respectively.
%For the VGGFace2 matcher on the whole dataset, the d-prime for the young, middle and old age range distributions is 6.596, 6.651, and 5.769, respectively.
%For the ArcFace matcher on the whole dataset, the d-prime for the young, middle and old age %range distributions is 8.952, 8.613, and 7.697, respectively.
As shown in Figure \ref{fig:hist}, in general, the d-prime values for younger and middle age ranges are more similar, and the d-prime for the old age range is noticeably worse.
% , which resulted in its worse performance.

In summary, the impostor distribution is the main cause for the old age range having a worse ROC curve than the other age ranges.
In contrast, the young age range's better ROC curves are explained mostly by its genuine distribution; its impostor distribution is worse than the middle age range.
% , but the genuine distribution outweighs that.
% These points are made more explicitly in the FMR and FNMR curves in the next section.

% \subsection{FMR and FNMR Curves}
% \label{sec:fmr_fnmr}
%Figure \ref{fig:fmr_fnmr} shows the FMR and FNMR curves corresponding to the distributions in Figure 3.  The FMR curves reinforce that, for any fixed decision threshold, the old age range has a worse (higher) FMR than the other two age ranges.
%The FMR for the young and middle age range are more similar to each other.
%The FNMR curves reinforce that, for any fixed decision threshold, the young age range has a better (lower) FNMR that the other two age ranges.
%The FNMRs for the middle and old age ranges are more similar to each other.

\subsection{Bootstrap Confidence Analysis}
\label{sec:confidence}

\begin{figure*}[t]
    \centering
    \begin{subfigure}[b]{0.82\textwidth}
        \centering
        \begin{subfigure}[b]{0.327\textwidth}
            \centering
            \includegraphics[width=1\columnwidth]{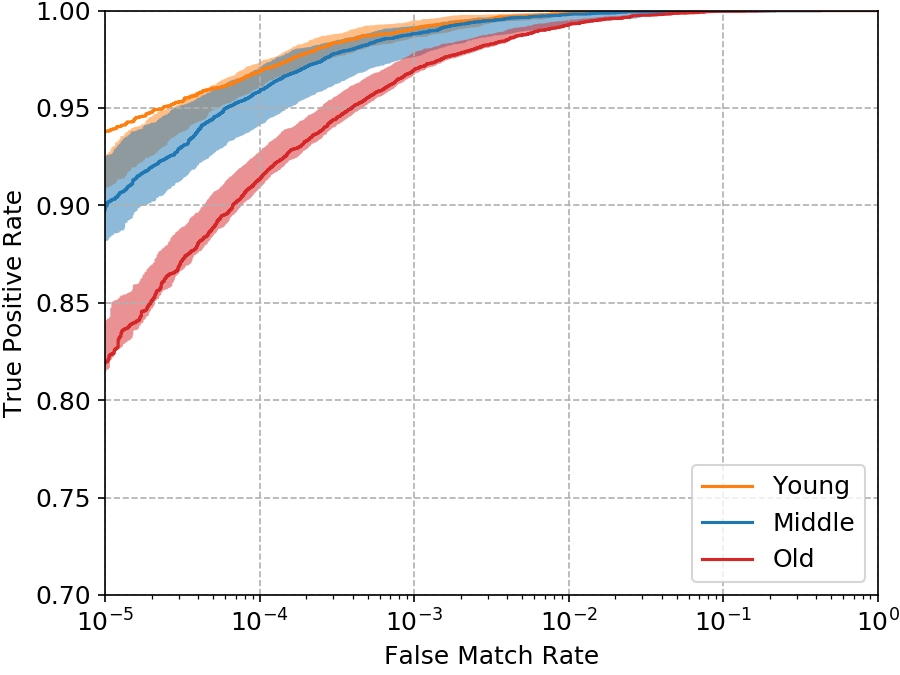}
            \caption{FaceNet}
            \vspace{-0.5em}
        \end{subfigure}
    %\end{subfigure}
    %\begin{subfigure}[b]{1\columnwidth}        
        \begin{subfigure}[b]{0.327\textwidth}
            \centering
            \includegraphics[width=1\columnwidth]{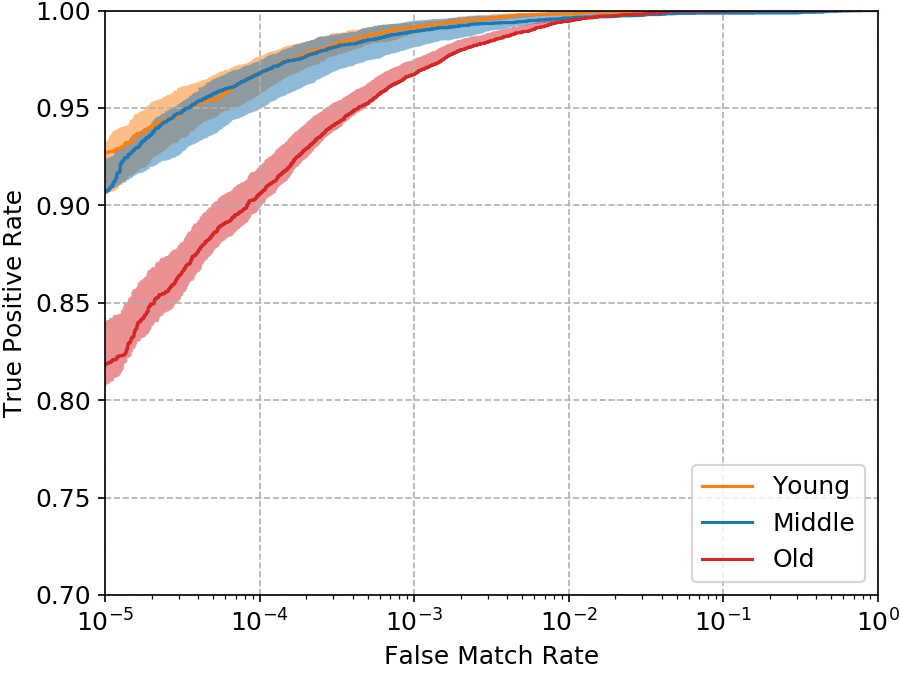}
            \caption{VGGFace2}
            \vspace{-0.5em}
        \end{subfigure}
        \hfill 
        \centering
        \begin{subfigure}[b]{0.327\textwidth}
            \centering
            \includegraphics[width=1\columnwidth]{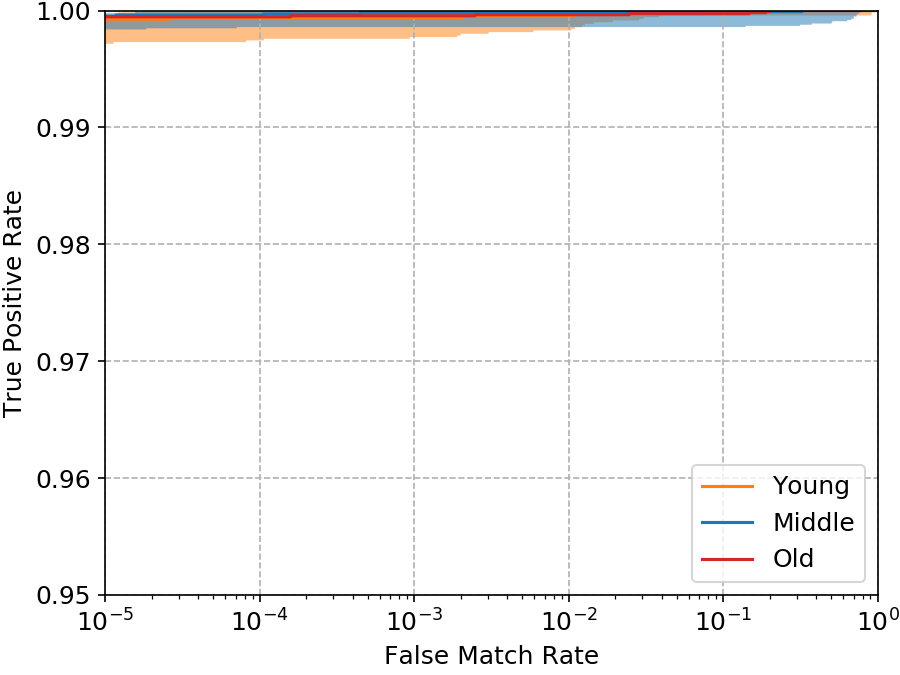}
            \caption{ArcFace}
            \vspace{-0.5em}
        \end{subfigure}
    \end{subfigure}
    \caption{ROC curves with 90\% confidence interval using 860 subjects randomly selected 100 times. ArcFace is displayed at a different scale for better visualization.}
    \label{fig:roc_confidence}
    \vspace{-1em}
\end{figure*}
To check whether our results might be an accident of the particular set of subjects and images, 
we randomly selected 80\% (860) of the subjects in the old group 100 times.
And to check whether our results might be partly due to the different number of subjects and images in the age ranges, 
% It is possible that the old age group's worse accuracy is due to a particular distribution of subjects.
% To validate that, using the whole dataset, 
% Also, someone could say that the young and middle age group better performance is due to them having many more subjects and images.
% To discard that possibility, 
we randomly selected 860 subjects in the young and middle age groups 100 times, so that they have the same number of subjects as the old age group.
We then computed 100 ROC curves for each age range and ordered them by the area under the curve (AUC) between a FMR of $10^{-5}$ and $10^{-3}$.
Figure \ref{fig:roc_confidence} shows the median ROC curve with a 90\% confidence interval using the ROC curves with 5-th and 95-th AUCs.

It is clear that generally the old group has a much worse ROC than the young and middle groups, even though all three age ranges have the same number of subjects.
Also, the ROC curves from Figure \ref{fig:roc} lay within the confidence intervals, indicating that the previous results are not the result of different subject/image quantity, or of an unusual distribution of subjects.
Finally, the middle age group shows a wider confidence interval, which overlaps with the young age confidence interval.

\begin{figure}[t]
    \begin{subfigure}[b]{1\columnwidth}
        \centering
        \begin{subfigure}[b]{.32\textwidth}
            \centering
            \includegraphics[width=1\columnwidth]{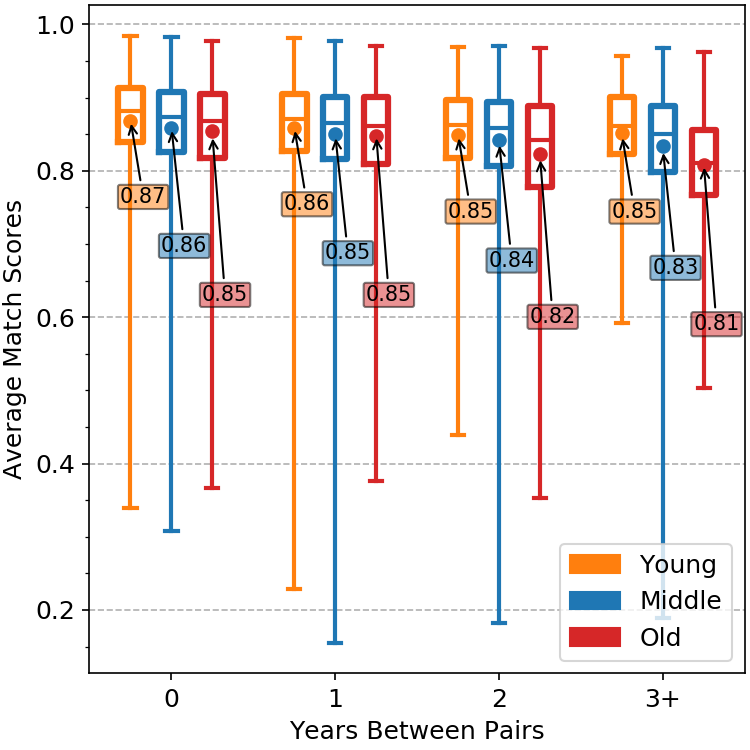}
        \end{subfigure}
        \hfill 
        \begin{subfigure}[b]{.32\textwidth}
            \centering
            \includegraphics[width=1\columnwidth]{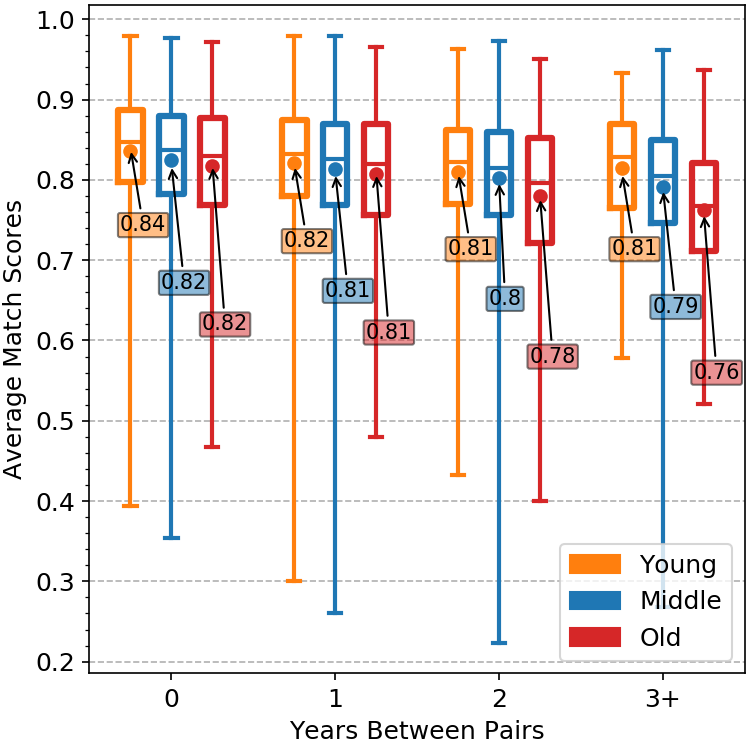}
        \end{subfigure}
    %\end{subfigure}
    \hfill
    %\begin{subfigure}[b]{0.32\textwidth}
        \begin{subfigure}[b]{.32\textwidth}
            \centering
            \includegraphics[width=1\columnwidth]{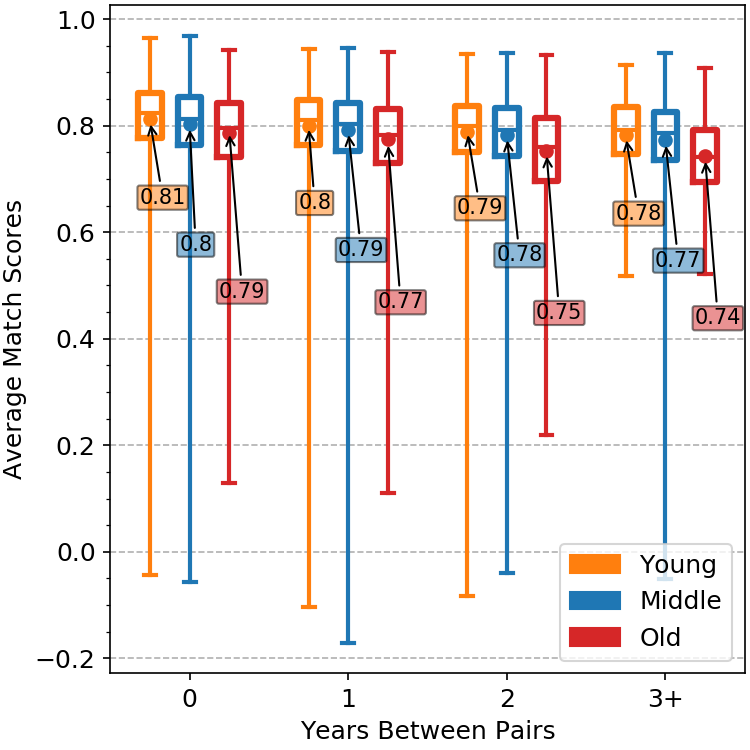}
        \end{subfigure}
        \hfill 
        \begin{subfigure}[b]{.32\textwidth}
            \centering
            \includegraphics[width=1\columnwidth]{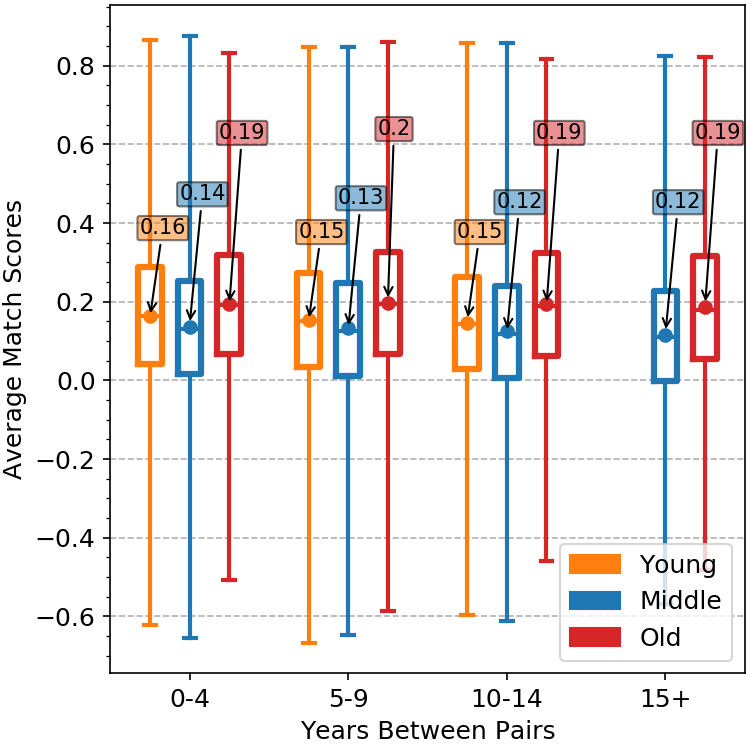}
            \caption{FaceNet}
            \vspace{-0.5em}
        \end{subfigure}
    %\end{subfigure}
    \hfill
    %\begin{subfigure}[b]{0.32\textwidth}
        \begin{subfigure}[b]{.32\textwidth}
            \centering
            \includegraphics[width=1\columnwidth]{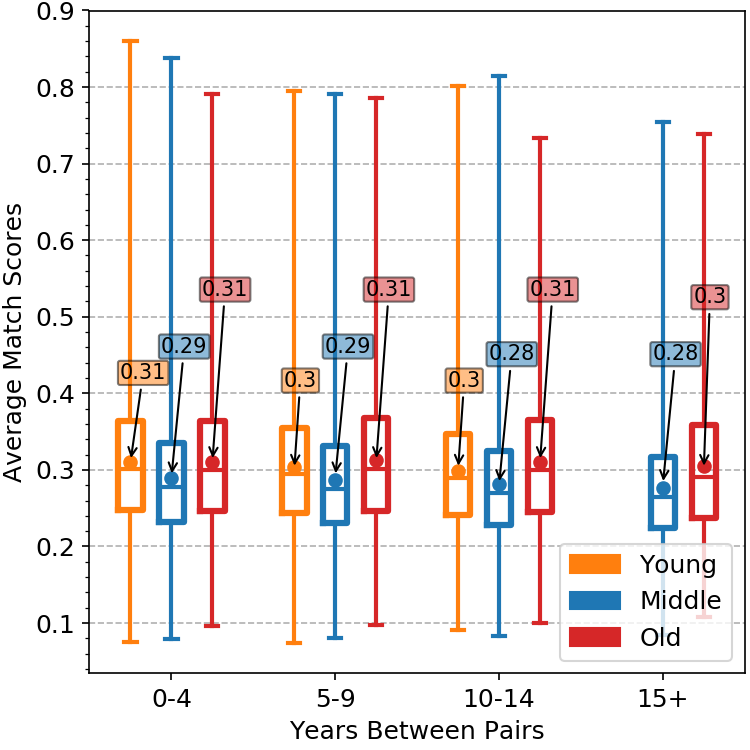}
            \caption{VGGFace2}
            \vspace{-0.5em}
        \end{subfigure}
        \hfill 
        \begin{subfigure}[b]{.32\textwidth}
            \centering
            \includegraphics[width=1\columnwidth]{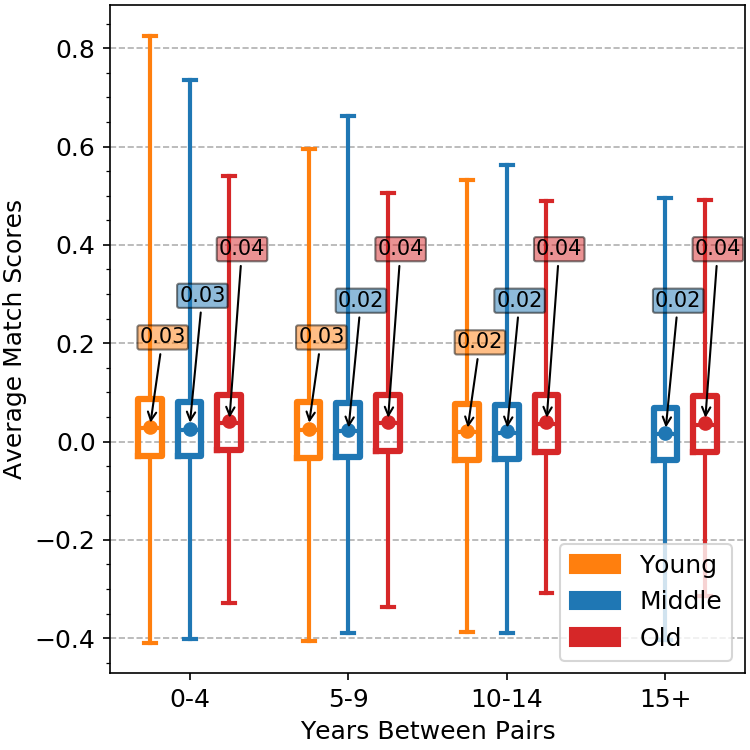}
            \caption{ArcFace}
            \vspace{-0.5em}
        \end{subfigure}
    \end{subfigure}
    \caption{Match scores with increasing elapsed time between authentic (top) and impostors (bottom) pairs for whole dataset.}
    \vspace{-1em}
    \label{fig:avg}
\end{figure}

\subsection{Influence of Elapsed Time}
\label{sec:elapsed_time}

Figure \ref{fig:avg} shows the average match score for impostor pairs and for genuine pairs, as a function of age difference between the person(s) in the two images.
For genuine pairs, this is age difference (time lapse) between images of the same person.
For impostor pairs, this is age difference between different persons.
The concept is likely most familiar for genuine matches.
The average match score between two images of the same person generally decreases with increase elapsed time.
A similar result holds for impostor matches.
Images of two different persons are on average more similar if the persons are the same age, and increased difference in ages generally decreases the average similarity.
The results are presented only for the whole dataset, as it has more images to break into separate bins of elapsed time.

For the genuine score plots, we see that the younger age range generally has the highest similarity scores and the older age range generally has the lowest similarity scores, for all time lapse bins and for all three matchers.
In effect, that data says that, on average, for the same length of time lapse, two images of the same older person look less similar than two images of the same younger person.
Also, we see a general trend of decreasing similarity score with increased elapsed time, which was expected. 

For the impostor score plots, 
% shown on the bottom of the figure,
the older age range generally has the worst (highest) average similarity scores and the middle age range generally has the best (lowest) average similarity scores.
For the young and middle age ranges, there is a small decrease in scores as a function of age increase between impostor pairs.
In the other hand, the old age range shows steady scores as the age difference between impostor pairs increase.

%For the young age range, the max number of years between persons in an impostor pair is 13, whereas it can be larger for the other two age ranges.
%However, the impostor distributions for the middle and old age ranges do not change significantly if all impostor pairs with more than 13 years age difference are dropped, and the comparison across age ranges does not change, as the impostor scores for the old group are consistently higher than both other groups, no matter the difference in age between the subjects.

Overall, we see no evidence that a difference in the distribution of time lapse between images across the different age ranges is a driving factor in the observed accuracy differences between the age ranges.

% On the other hand, in one matcher, the young age range scores decreases until 2 years and then increase again, which could be correlated to the small number of matching pairs with more than 3 years lapsed time, which is only 195.
% Moreover, the young age range has the best overall scores, followed by the middle age range, and then the old age range.
% In general, all three groups have similar scores across pairs with different lapsed time between them.

% Moving to the impostor distribution, shown on the bottom, both middle and young age range show decreasing trends with the increase in age difference between pairs, which again, was expected. 
% However, the old age range shows steady scores, or even increasing scores as the age difference between the pairs increase.
% In contrast to the overall average scores for genuine scores, the impostor scores for the old age range show overall higher values. 
% Also, the middle age range shows lower scores than the young age range.

\subsection{Comparison with Pre Deep CNN Algorithm}
\label{sec:pre_cnn_results}
The best-known previous works comparing accuracy across different age ranges date to before the wave of deep CNN matchers; e.g. \cite{Klare2012}.
Also, previous works use different datasets from the one that we use.
Therefore, our results being different from previous results could potentially be due to (a) newer deep learning matchers having different properties than older matchers, and / or (b) our dataset  being different somehow from those used in previous work.
The PCSO dataset analyzed in \cite{Klare2012} is no longer available.
However, we re-implemented the Local Binary Patterns (LBP) matcher as described in \cite{Klare2012} and used it to analyze our dataset.

%Our re-implemented LBP matcher has four steps. 
%First, we resize the aligned face images to 108x108. 
%Second, we normalize the illumination with the method described in \cite{tan_triggs}.
%Third, the image is divided into 9x9 regions with 12x12 pixels each, which are then processed by the LBP feature extractor \cite{lbp_face} with radii 1 and 2. 
%Lastly, the histograms are concatenated, generating feature vectors of 9558-d, which are then matched using Chi-Square distance.

\begin{figure}[t]
    \centering
    \begin{subfigure}[b]{1\columnwidth}
        \centering
        \begin{subfigure}[b]{0.32\columnwidth}
            \centering
            \includegraphics[width=1\columnwidth]{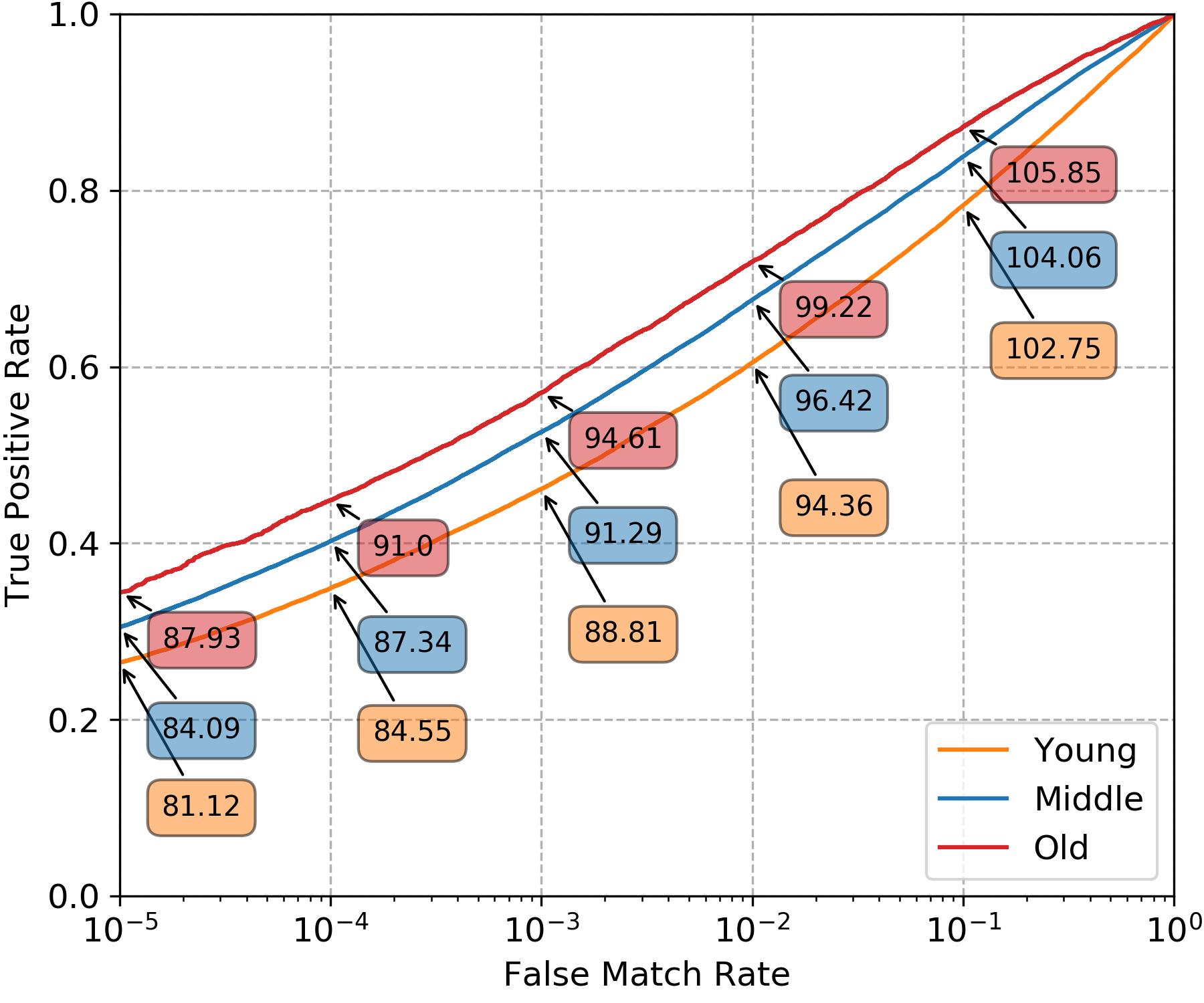}
            \caption{Whole dataset}
            \vspace{-0.5em}
        \end{subfigure}
    %\end{subfigure}
    %\begin{subfigure}[b]{1\columnwidth}        
        \begin{subfigure}[b]{0.32\columnwidth}
            \centering
            \includegraphics[width=1\columnwidth]{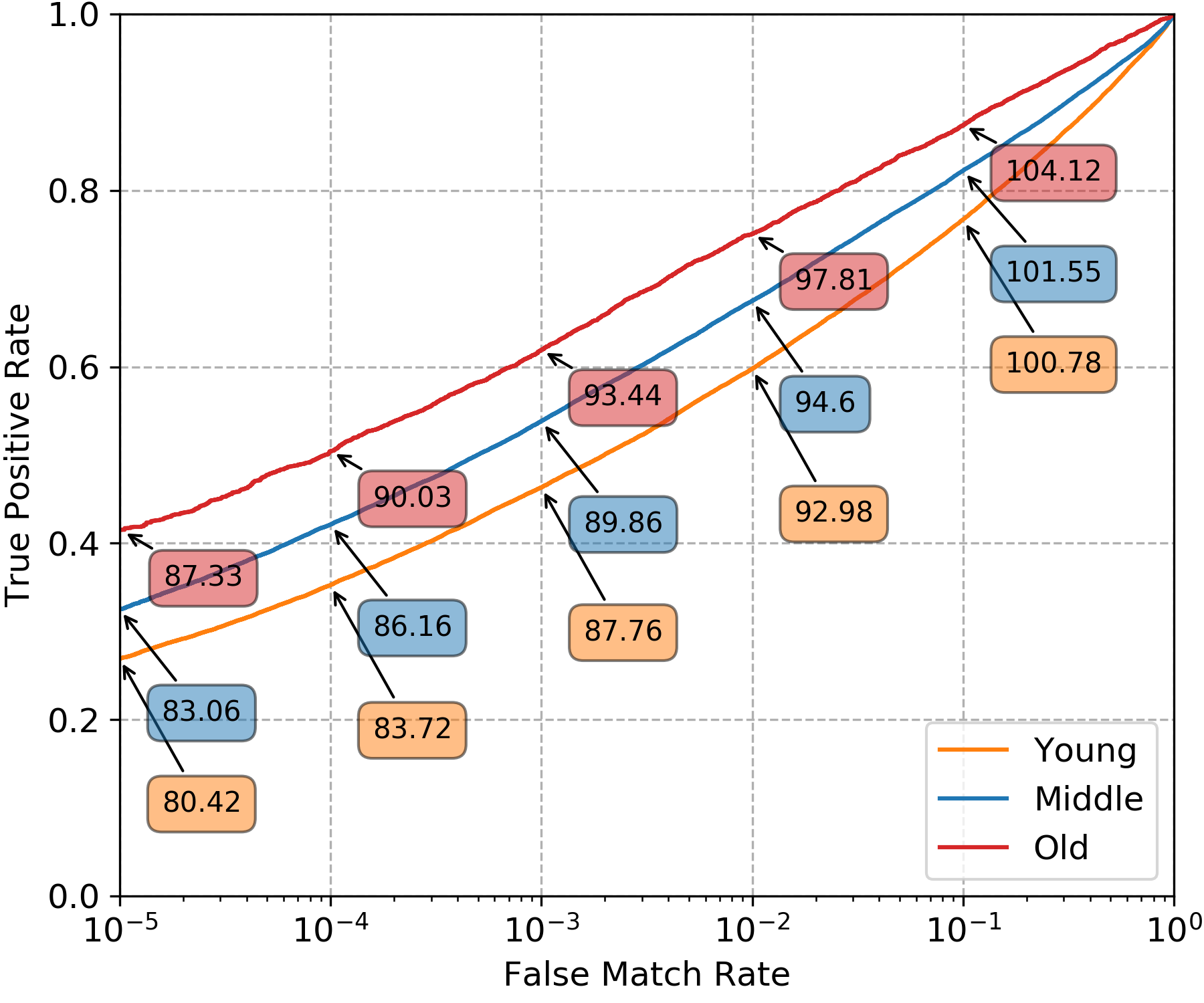}
            \caption{AA males}
            \vspace{-0.5em}
        \end{subfigure}
        \hfill 
        \centering
        \begin{subfigure}[b]{0.32\columnwidth}
            \centering
            \includegraphics[width=1\columnwidth]{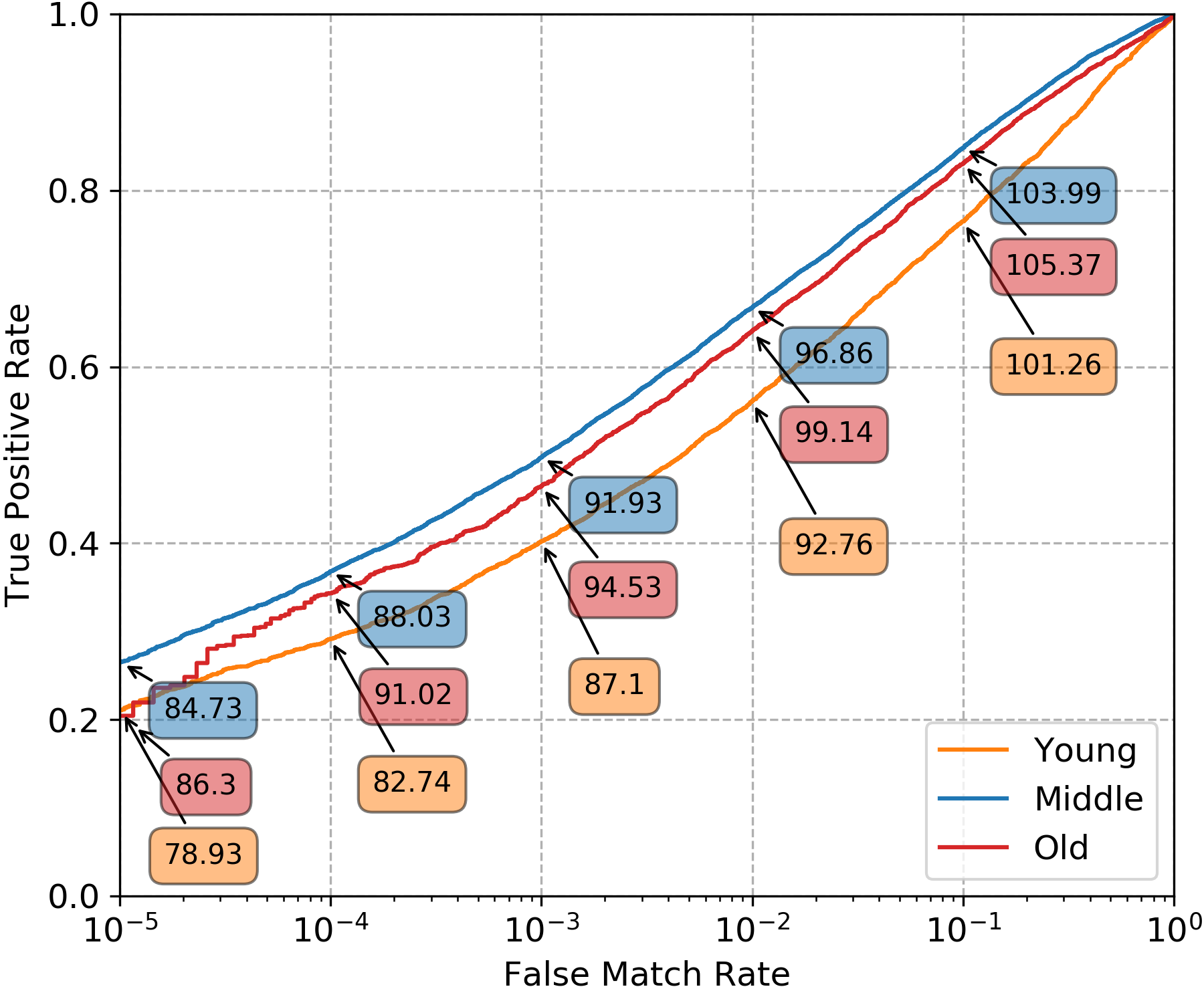}
            \caption{C males}
            \vspace{-0.5em}
        \end{subfigure}
    \end{subfigure}
    \caption{ROC curves for LBP matcher for whole dataset, African American (AA) males, and Caucasian (C) males. Annotated values correspond to thresholds used for the correspondent FMR.}
    \label{fig:roc_lbp}
    \vspace{-1em}
\end{figure}

Figure \ref{fig:roc_lbp} shows ROC curves for the LBP matcher.
We can see that the trend across age ranges for the LBP matcher is the opposite of the trend for the deep CNN matchers.
For the LBP matcher on our dataset, the young age range consistently has the worst performance, and the old age range has the best performance in two of the subsets.
Thus the results of the LBP matcher on our dataset agree qualitatively with the results of the LBP matcher on the PCSO dataset as reported in \cite{Klare2012}.
This indicates that our results differ from those of past studies due to modern deep CNN matchers operating differently from older matchers and not because of some difference in the datasets.

\subsection{Training Data Analysis}
\label{sec:training_data}

Using an CNN age predictor \cite{insightface}, we predicted the subject age in the images in the three datasets that were used in training the three CNN matchers.
The age predictor mean absolute error (MAE) on the whole MORPH dataset is 8.55, and is 4.1 on the validation set.
Because the age predictor has MAE around 4 on the validation test, we classified faces predicted as 34 or less years as young, 54 or less as middle age, and 55 or more as old.
%We chose 34 and 54 instead of 30 and 50 because .
Table \ref{tab:training_ages} clearly shows that all three datasets are imbalanced, with the middle age range having the most representation and the older age range having the least.
This shows that the better accuracy observed for the young age range is not directly caused by the fraction of young faces in the training data, as the middle age range represents a larger fraction of the training data. 
It also demonstrates that the much better and more similar accuracy of ArcFace across age groups is not due to the training data being more balanced than for the other matchers, but mostly because of its better loss function.

\begin{table}[t]
    \centering
    \setlength\tabcolsep{3.5pt}
    \small
    \ra{1.1}
    \begin{tabular}{l|l|l|l}
        & \multicolumn{3}{c}{\textbf{Number of Images}} \\
        \textbf{Dataset} & \multicolumn{1}{c|}{\textbf{Young}} & \multicolumn{1}{c|}{\textbf{Middle}} & \multicolumn{1}{c}{\textbf{Old}} \\ \hline
        \textbf{VGGFace2} & 1.01M (32.5\%) & 1.74M (55.9\%) & 0.36M (11.6\%)\\
        \textbf{MS1M} & 2.97M (35.2\%) & 4.49M (53.1\%) & 0.99M (11.7\%)\\
        \textbf{MS1M V2}& 1.66M (28.6\%) & 3.26M (56.1\%) & 0.89M (15.3\%) 
    \end{tabular}
    \caption{Number of images and ratio of young, middle and old age groups in the training datasets.}
    \label{tab:training_ages}
    \vspace{-1.5em}
\end{table}

% \subsubsection{Fine-tuning on Age Groups}
To further investigate the effects of the subjects' ages in the training data, we fine-tuned the two higher-accuracy matchers using subjects in a specific age range.
The age range with the smaller amount of images and subjects is the old range, which was selected as the starting point for each training group preparation.
First, for each age range, we removed all subjects that have less than 50 images, which yields 2,009 subjects with 308,640 images for VGGFace2, and 5,033 subjects with 348,742 images for MS1M V2.
Then, we randomly selected the same number of subjects and images for the young and middle age groups, to make training subsets balanced on number of subjects and images.

\begin{table*}[t]
    \small
    \setlength\tabcolsep{4pt}
    \centering
    \ra{1.1}
    \begin{tabular}{l|rrrr|rrrr|rrrr}
        \multicolumn{1}{l|}{}& \multicolumn{4}{c|}{\textbf{Whole Dataset}} & \multicolumn{4}{c|}{\textbf{African American Males}}& \multicolumn{4}{c}{\textbf{Caucasian Males}}\\
        \textbf{Matcher@Trainining subset} & \multicolumn{1}{c}{\textbf{Young}} & \multicolumn{1}{c}{\textbf{Middle}} & \multicolumn{1}{c}{\textbf{Old}} & \multicolumn{1}{c|}{\textbf{Diff.}} & \multicolumn{1}{c}{\textbf{Young}} & \multicolumn{1}{c}{\textbf{Middle}} & \multicolumn{1}{c}{\textbf{Old}} & \multicolumn{1}{c|}{\textbf{Diff.}} & \multicolumn{1}{c}{\textbf{Young}} & \multicolumn{1}{c}{\textbf{Middle}} & \multicolumn{1}{c}{\textbf{Old}} & \multicolumn{1}{c}{\textbf{Diff.}} \\ \hline
        VGGFace2@MSM1V2\_Young & 92.32& 87.3& 81.39& 10.93 & 94.43& 91.05 & 87.06& 7.37& 94.33& 90.96 & 84.18& 10.15\\
        VGGFace2@MSM1V2\_Middle& \textbf{93.24} & \textbf{90.44}& \textbf{85.64} & \textbf{7.6}& \textbf{95.41} & \textbf{93.2} & \textbf{90.16} & 5.25& \textbf{94.82} & \textbf{93.44}& 85.76& 9.06 \\
        VGGFace2@MSM1V2\_Old & 88.54& 86.67 & 80.75& 7.79& 91.02& 90.41 & 88.45& \textbf{2.57} & 92.93& 91.75 & \textbf{89.17} & \textbf{3.76}\\ \hline
        ArcFace@VGGFace2\_Young& 98.39& 98.47 & 95.08& 3.39& 99.12& 99.14 & 97.32& 1.82& 98.69& 99.31 & 97.41& 1.9\\
        ArcFace@VGGFace2\_Middle & 98.19& 98.33 & 95.77& 2.56& 98.93& 98.93 & 97.41& 1.52& 98.41& 98.93 & 97.97& 0.96 \\
        ArcFace@VGGFace2\_Old& \textbf{99.37} & \textbf{99.37}& \textbf{99.27} & \textbf{0.1}& \textbf{99.47} & \textbf{99.51}& \textbf{99.35} & \textbf{0.16} & \textbf{99.55} & \textbf{99.94}& \textbf{99.81} & \textbf{0.39}\\ \hline \hline
        
        \textit{VGGFace2 Baseline} & \textit{96.94} & \textit{96.94} & \textit{90.51} & \textit{6.43} & \textit{97.49} & \textit{97.02} & \textit{94.04} & \textit{3.45} & \textit{96.3}  & \textit{95.71} & \textit{90.81} & \textit{5.49} \\
        \textit{ArcFace Baseline}  & \textit{99.89} & \textit{99.95} & \textit{99.95} & \textit{0.06} & \textit{99.89} & \textit{99.95} & \textit{99.95} & \textit{0.06} & \textit{99.91} & \textit{99.95} & \textit{99.96} & \textit{0.05} \\
        
        \hline \hline
        VGGFace2@MS1MV2\_All\_ft & \textbf{95.84} & \textbf{93.23}& \textbf{89}& 6.84& \textbf{96.83} & \textbf{95.15}& \textbf{93.51} & 3.32& \textbf{95.92} & \textbf{94.08}& \textbf{90.69} & 5.23 \\
        VGGFace2@MS1MV2\_All & 77.01& 75.53 & 69.05& 7.96& 77.35& 76.66 & 74.94& 2.41& 76.82& 74.04 & 73.03& \textbf{3.79}\\
        VGGFace2@VGGFace2\_All & 71.21& 70.99 & 67.85& \textbf{3.36} & 72.15& 73.18 & 74.19& \textbf{2.04} & 76.28& 75.28 & 70.3 & 5.98 \\ \hline
        ArcFace@VGGFace2\_All\_ft& 98.27& 98.56 & 97.16& 1.4 & 98.74& 99& 98.29& 0.45& 98.26& 99.06 & 98.51& \textbf{0.8} \\
        ArcFace@VGGFace2\_All& 95.82& 96.72 & 95.02& 1.7 & 96.55& 97.69 & 96.44& 1.25& 95.66& 97.36 & 96.82& 1.7\\
        ArcFace@MS1MV2\_All& \textbf{98.85} & \textbf{99.21}& \textbf{99.27} & \textbf{0.42} & \textbf{99.05} & \textbf{99.39}& \textbf{99.12} & \textbf{0.34} & \textbf{98.2}& \textbf{99.14}& \textbf{99.68} & 1.48
    \end{tabular}
    \caption{True positive rates (\%) with a false match rate of $10^{-4}$ and difference (best $-$ worst) for models fine-tuned with different age group subsets (top half), and fine-tuned (ft) or trained from scratch on all age groups combined (bottom half). The baseline results are shown in the middle of the table.} 
    \label{tab:fine_tuning}
    \vspace{-1.5em}
\end{table*}

Using each age group separately, we fine-tuned the VGGFace2 matcher using data from the MS1M V2 dataset, %using its original loss (softmax), 
and the ArcFace matcher using data from the VGGFace2 dataset, so that we do not re-use images from a matcher's original training data as fine-tuning data.
%also with its original loss (additive angle margin).
% The learning rate starts at 0.001, and we use mini-batch size of 256 images.
% For the VGGFace2 matcher, early stop is applied if the training accuracy does not increase at least 0.1\% for two consecutive epochs.
%For the ArcFace matcher, each traininig runs for 10K iterations.
Table \ref{tab:fine_tuning} (upper half) shows the TPR at a FMR of $10^{-4}$ for each fine-tuned model.
We observe that the performance dropped for all the fine-tuned models compared to the original results (shown in middle), which may be expected as there are many fewer images and subjects, thus less generalization capacity.
In general, no matter the age range used for fine-tuning, the old age group has the worst accuracy.
Contrary to what might be expected, the best results for each age group are not correlated to fine-tuning on the same age range, e.g. the young age range had better accuracy when the matcher was fine-tuned either on a middle age group (VGGFace2) or an old age group (ArcFace).
For the VGGFace2 matcher, in general, the best accuracy for the three age groups were with the fine-tuning on the middle age group, with d-prime values in the whole dataset of 5.273, 5.518 and 5.082 for young, middle, and old age groups, respectively.
ArcFace achieved better accuracy for all age groups when fine-tuned on an old age group, with d-prime values in the whole dataset of 7.919, 7.908, 7.247 for young, middle, and old age groups, respectively.
Moreover, in five out of six results, the difference between the best and worst group was lower when the matcher was fine-tuned using an old age group.
When fine-tuning with a young age group, the difference between the best and worst age group is much higher for both matchers, which indicates that middle to old faces are better for a more uniform and higher accuracy between age groups.

% \subsubsection{Training on Age Balanced Dataset}
It is possible that the worse accuracy for the old age range across the fine-tunings is due to some bias in the initial training.
To explore that, we combined the three age subsets together, creating age balanced datasets.
%with a total of 925,920 images in VGGFace2, and 1,046,226 images in MS1M V2.
Then, we fine-tuned and trained from scratch the VGGFace2 and ArcFace matchers on the age balanced datasets.
%with the difference that the ArcFace runs for 30K iterations now.
%train both matchers from scratch using the same 
%The learning rate starts at 0.1 and is performed with mini-batch size of 256 images.
%For VGGFace2 matcher, the learning rate is reduced twice it the loss plateus, and early stop is applied when the training accuracy does not increase for two epochs.
%For the ArcFace matcher, is reduced at 50K and 70K iterations, and the training ends at 100K iterations.

Table \ref{tab:fine_tuning} (lower half) shows results for the matchers fine-tuned and trained from scratch on the age balanced datasets.
Only one out of six results shows closer performance between the worst and best group for the fine-tuning results.
For all other five results, a more similar performance was achieved training from scratch.
With the VGGFace2 matcher, the accuracy with the fine-tuning is much higher than when trained from scratch, but the performance is still lower for the old group, and the d-prime values for the whole dataset are 4.563, 4.661, and 4.315 for young, middle, and old age groups, respectively.
On the other hand, ArcFace achieved better performance when training from scratch on the MS1M V2 dataset, with a higher accuracy for the old group compared to the young, but still with a lower d-prime value for the old group, which is 6.876, compared to 7.06 for the young group, and 7.367 for middle group.
Moreover, ArcFace training and fine-tuning on VGGFace2 dataset also show worse results for the old group compared to the other two groups.
Finally, the results show that even when the dataset is age balanced, in general, the old group has lower performance than young or middle.

\section{Conclusions and Discussion}
\label{sec:conclusion}

\textbf{Younger people are easier, older people are harder.}
Previous works have found that people in the old age range are easier to recognize (have higher accuracy) than people in the young age range.
However, we find the opposite.
For two of the three modern deep CNN face matchers used in our work, the old age range has a noticeably worse ROC curve than the young age range, and the other matcher, which has much higher accuracy for all age ranges, shows similar accuracy because all age groups are basically ``at ceiling''.
The bootstrap results also confirm that this pattern is not due to number of subjects or particular age group distributions of the dataset.

The genuine distributions for the three age ranges seem more similar than their impostor distributions.
Therefore it seems that the impostor distribution for the old age range is driving its worse ROC. 
Also, the d-prime values show how much worse the separation between the old group distributions is compared to young and middle.

\textbf{The new deep CNN face matchers are different.}
The different pattern of accuracy across age ranges in our results could, in principle, be due to a difference in testing datasets or to modern face matchers having different properties.
Cao et al. \cite{vggface2} describe an experiment involving two age groups, ``young'' (less than 34 years) and ``mature'' (34 years or more), in which they report that ``mature'' is recognized more accurately than ``young'' with the VGGFace2 matcher.
This result appears to align better with the traditional expectation for young/old accuracy, than with our result.
However, this result is based just on 100 subjects.
Also, it uses 2 ``templates'' of 5 images each, for each age range for each subject, for a total of just 2,000 images.
Examining the 20 images in the templates for the first subject number of the 100 subjects, we found that there are duplicate images, incorrect identity labels, incorrect age category labels, as well as images with extreme pose variation.
(These 20 images are shown in the Supplemental Materials.)
Given these issues with the images and meta-data, combined with the small number of subjects and images, the young/mature result found by Cao et al. \cite{vggface2} simply may not be reliable.

Lu et al. \cite{Lu2018} report results across five age ranges (that were predicted using crowd sourcing) with deep CNN face matchers.
Their results show increasing accuracy until the age of 50, then a drop in accuracy,
which agrees with our results on the old age group, but disagrees on the young age group.
However, as the dataset used is unconstrained, there are many factors that can be affecting one age group more than other, e.g., pose, illumination, and facial expression.
So, it is possible that the lower performance of subjects in the middle 20s is not related to their ages, but to external factors.

As far as we know, MORPH \cite{morph} is the only currently available dataset with recorded age meta-data, enough subjects across ages, and consistent quality, so, we cannot reproduce the experiments on another dataset.
However, we re-implemented a pre deep CNN face matcher (LBP) as described in \cite{Klare2012} and ran it on our dataset and obtained results similar to those in \cite{Klare2012}.
This indicates that the pattern of results in our work is due to newer deep CNN matchers, rather than a property of the dataset.  

Deep CNN face matching technology seems to have changed the default expectation for differences in accuracy across age ranges.
The datasets that current deep CNN matchers have been trained on do not have a large fraction of older subjects.
For instance, all the datasets used to train the matchers tested in this work, have less than 16\% of the images with a subject older than 50 years.
It would be possible that models trained on a dataset with a more balanced age distribution would have a similar performance across age groups.
However, the highest-accuracy matcher, ArcFace, achieves more similar results across age groups, and has a very imbalanced training dataset.
To further investigate the training data effect, we fine-tuned two matchers on different age groups, and the results are generally worse for the old age group, no matter the age subset fine-tuned on.
Moreover, fine-tuning or training from scratch on an age balanced dataset does not achieve the same performance across ages, showing in general worse results for the older age group and better results for the younger or middle groups.
Therefore, we conclude that balanced training data is not the simple answer to achieve same performance across ages; rather, better loss functions are desirable.

\textbf{Time-lapse changes impostor as well as genuine.}
The “template aging” effect is well known - 
increasing time lapse between images in an genuine pair results in a lower similarity score.
We show that an increased difference in the age between two persons in an impostor pair also results in a lower similarity score.
This is perhaps intuitive – images of two different persons of the same age are likely to look more alike than images of two different persons with large age difference.
This result appears consistent with results presented by Grother \cite{grother}.
However, our result for the old age range is atypical on this point; increased age difference between persons in an impostor pair has a less predictable effect, staying about the same or even increasing slightly in one case. 
We do not yet have any confident speculation for the cause of this effect. 
% There are a relatively small number of impostor pairs in the old group with a large age difference, and so statistical fluctuation may play a role.

%\vfill
%\pagebreak

{\small
\bibliographystyle{ieee}
\bibliography{main}
}

\vfill
\pagebreak
\newgeometry{top=0.63in, left=0.63in, bottom=0.63in, right=0.63in}

\onecolumn

% \appendix

% \subsection{
\begin{center}
{\bf Supplementary Material - Illustration of Issues with VGGFace2 Age Templates}\\
\end{center}
% }

\begin{figure}[h]
    \begin{subfigure}[b]{1\columnwidth}
        \centering
        \begin{subfigure}[b]{.17\textwidth}
            \centering
            \includegraphics[width=1\columnwidth]{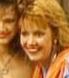}
            \renewcommand\thesubfigure{\alph{subfigure}1}
            \vspace{-1.5em}
            \caption{Anne Schedeen}
        \end{subfigure}
        \hfill 
        \begin{subfigure}[b]{.17\textwidth}
            \centering
            \includegraphics[width=1\columnwidth]{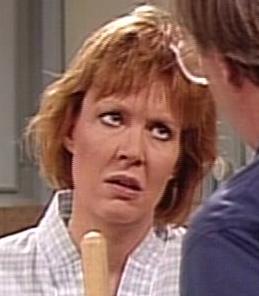}
            \addtocounter{subfigure}{-1}
            \renewcommand\thesubfigure{\alph{subfigure}2}
            \vspace{-1.5em}
            \caption{Anne Schedeen}
        \end{subfigure}
        \hfill
        \begin{subfigure}[b]{.17\textwidth}
            \centering
            \includegraphics[width=1\columnwidth]{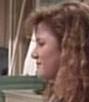}
            \addtocounter{subfigure}{-1}
            \renewcommand\thesubfigure{\alph{subfigure}3}
            \vspace{-1.5em}
            \caption{\textit{Unconfirmed}}
        \end{subfigure}
        \hfill 
        \begin{subfigure}[b]{.17\textwidth}
            \centering
            \includegraphics[width=1\columnwidth]{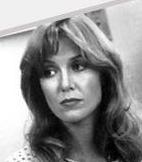}
            \addtocounter{subfigure}{-1}
            \renewcommand\thesubfigure{\alph{subfigure}4}
            \vspace{-1.5em}
            \caption{Anne Schedeen}
        \end{subfigure}
        \hfill
        \begin{subfigure}[b]{.17\textwidth}
            \centering
            \includegraphics[width=1\columnwidth]{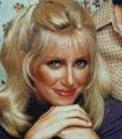}
            \addtocounter{subfigure}{-1}
            \renewcommand\thesubfigure{\alph{subfigure}5}
            \vspace{-1.5em}
            \caption{{\color{red}{Suzanne Sommers}}}
        \end{subfigure}
        %\addtocounter{subfigure}{-1}
        %\caption{Young Template 1}
        \vspace{0.5em}
    \end{subfigure}
    \begin{subfigure}[b]{1\columnwidth}
        \centering
        \begin{subfigure}[b]{.17\textwidth}
            \centering
            \includegraphics[width=1\columnwidth]{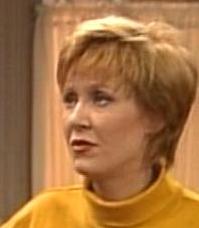}
            \renewcommand\thesubfigure{\alph{subfigure}1}
            \vspace{-1.5em}
            \caption{Anne Schedeen}
        \end{subfigure}
        \hfill 
        \begin{subfigure}[b]{.17\textwidth}
            \centering
            \includegraphics[width=1\columnwidth]{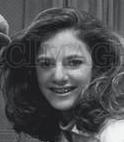}
            \addtocounter{subfigure}{-1}
            \renewcommand\thesubfigure{\alph{subfigure}2}
            \vspace{-1.5em}
            \caption{{\color{red}{Andrea Elson}}}
        \end{subfigure}
        \hfill
        \begin{subfigure}[b]{.155\textwidth}
            \centering
            \includegraphics[width=1\columnwidth]{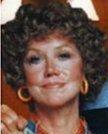}
            \addtocounter{subfigure}{-1}
            \renewcommand\thesubfigure{\alph{subfigure}3}
            \vspace{-1.5em}
            \caption{{\color{red}{Audra Lindley}}}
        \end{subfigure}
        \hfill 
        \begin{subfigure}[b]{.17\textwidth}
            \centering
            \includegraphics[width=1\columnwidth]{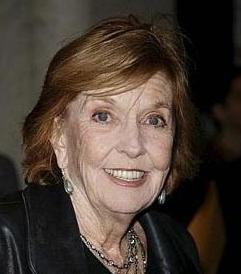}
            \addtocounter{subfigure}{-1}
            \renewcommand\thesubfigure{\alph{subfigure}4}
            \vspace{-1.5em}
            \caption{{\color{red}{Anne Meara}}}
        \end{subfigure}
        \hfill
        \begin{subfigure}[b]{.17\textwidth}
            \centering
            \includegraphics[width=1\columnwidth]{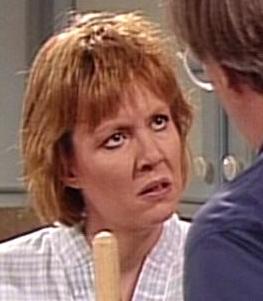}
            \addtocounter{subfigure}{-1}
            \renewcommand\thesubfigure{\alph{subfigure}5}
            \vspace{-1.5em}
            \caption{Anne Schedeen}
        \end{subfigure}
        %\addtocounter{subfigure}{-1}
        %\caption{Young Template 2}
        \vspace{0.5em}
    \end{subfigure}
    
    \begin{subfigure}[b]{1\columnwidth}
        \centering
        \begin{subfigure}[b]{.17\textwidth}
            \centering
            \includegraphics[width=1\columnwidth]{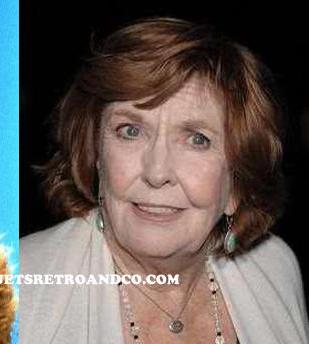}
            \renewcommand\thesubfigure{\alph{subfigure}1}
            \vspace{-1.5em}
            \caption{{\color{red}{Anne Meara}}}
        \end{subfigure}
        \hfill 
        \begin{subfigure}[b]{.17\textwidth}
            \centering
            \includegraphics[width=1\columnwidth]{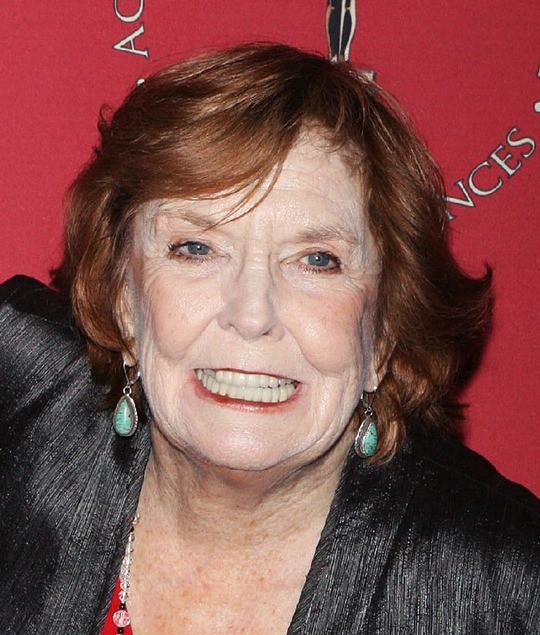}
            \addtocounter{subfigure}{-1}
            \renewcommand\thesubfigure{\alph{subfigure}2}
            \vspace{-1.5em}
            \caption{{\color{red}{Anne Meara}}}
        \end{subfigure}
        \hfill
        \begin{subfigure}[b]{.175\textwidth}
            \centering
            \includegraphics[width=1\columnwidth]{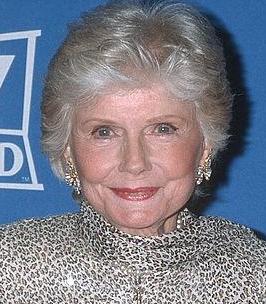}
            \addtocounter{subfigure}{-1}
            \renewcommand\thesubfigure{\alph{subfigure}3}
            \vspace{-1.5em}
            \caption{{\color{red}{Barbara Billingsley}}}
        \end{subfigure}
        \hfill 
        \begin{subfigure}[b]{.17\textwidth}
            \centering
            \includegraphics[width=1\columnwidth]{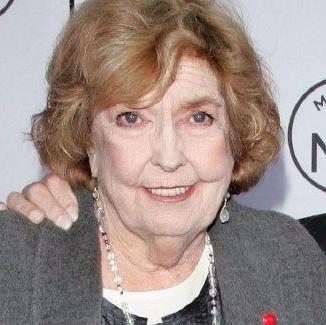}
            \addtocounter{subfigure}{-1}
            \renewcommand\thesubfigure{\alph{subfigure}4}
            \vspace{-1.5em}
            \caption{{\color{red}{Anne Meara}}}
        \end{subfigure}
        \hfill
        \begin{subfigure}[b]{.17\textwidth}
            \centering
            \includegraphics[width=1\columnwidth]{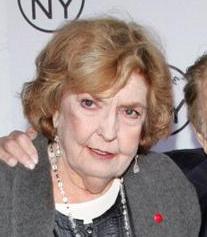}
            \addtocounter{subfigure}{-1}
            \renewcommand\thesubfigure{\alph{subfigure}5}
            \vspace{-1.5em}
            \caption{{\color{red}{Anne Meara}}}
        \end{subfigure}
        %\addtocounter{subfigure}{-1}
        %\caption{Mature Template 1}
        \vspace{0.5em}
    \end{subfigure}
    
    \begin{subfigure}[b]{1\columnwidth}
        \centering
        \begin{subfigure}[b]{.17\textwidth}
            \centering
            \includegraphics[width=1\columnwidth]{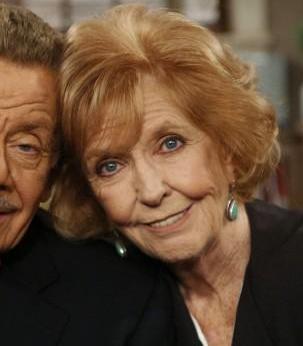}
            \renewcommand\thesubfigure{\alph{subfigure}1}
            \vspace{-1.5em}
            \caption{{\color{red}{Anne Meara}}}
        \end{subfigure}
        \hfill 
        \begin{subfigure}[b]{.17\textwidth}
            \centering
            \includegraphics[width=1\columnwidth]{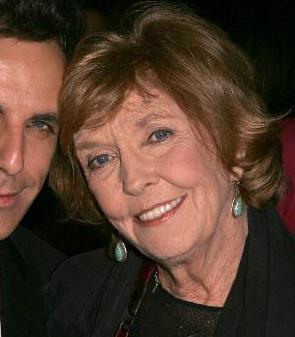}
            \addtocounter{subfigure}{-1}
            \renewcommand\thesubfigure{\alph{subfigure}2}
            \vspace{-1.5em}
            \caption{{\color{red}{Anne Meara}}}
        \end{subfigure}
        \hfill
        \begin{subfigure}[b]{.175\textwidth}
            \centering
            \includegraphics[width=1\columnwidth]{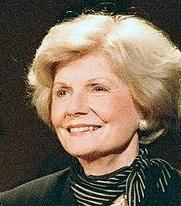}
            \addtocounter{subfigure}{-1}
            \renewcommand\thesubfigure{\alph{subfigure}3}
            \vspace{-1.5em}
            \caption{{\color{red}{Barbara Billingsley}}}
        \end{subfigure}
        \hfill 
        \begin{subfigure}[b]{.17\textwidth}
            \centering
            \includegraphics[width=1\columnwidth]{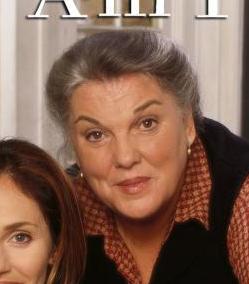}
            \addtocounter{subfigure}{-1}
            \renewcommand\thesubfigure{\alph{subfigure}4}
            \vspace{-1.5em}
            \caption{{\color{red}{Tyne Daly}}}
        \end{subfigure}
        \hfill
        \begin{subfigure}[b]{.17\textwidth}
            \centering
            \includegraphics[width=1\columnwidth]{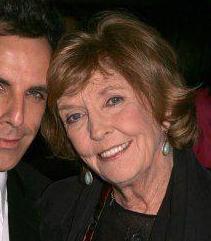}
            \addtocounter{subfigure}{-1}
            \renewcommand\thesubfigure{\alph{subfigure}5}
            \vspace{-1.5em}
            \caption{{\color{red}{Anne Meara}}}
        \end{subfigure}
        %\addtocounter{subfigure}{-1}
        %\caption{Mature Template 2}
    \end{subfigure}
    \caption{First subject of VGGFace2 Age Templates (Anne Schedeen - n000654) with {\color{red}{incorrect}} and correct labels. Young template \#1 and \#2 (first two rows), and mature template \#1 and \#2 (last two rows).}
    \label{fig:age_templates}
\end{figure}

This supplementary material is based on the VGGFace2 documentation\footnote{\url{http://www.robots.ox.ac.uk/~vgg/data/vgg_face2/meta_infor.html}}.
Based on the  AgeTemplates info, the first subject number in the age templates is n000654.
Based on the IdentityMeta info, subject n000654 is an actress named Anne Schedeen.
Figure \ref{fig:age_templates} shows the 20 images in the age templates for n000654 [6], along with actual identities as best we have been able to determine them.  
Among the 20 images, only 5 or 6 are of Anne Schedeen. 
% We could not confirm the identity of three images, however two seem to belong to her.
The person seen most in these templates is Anne Meara, co-star of Schedeen in "ALF".
% , in 7 of the 10  images for the "mature" templates.
Matching between the two mature templates would likely generate a higher score than matching between the younger templates, as there is a smaller number of subjects in the mature templates.

While we have only investigated in detail the first subject in the age templates, the problems are not unique to this subject.
We conclude that results presented for mature and young templates by Cao et al. [6] may not be reliable, and that a cleaning procedure may not leave enough images for a template style analysis.

\end{document}